\documentclass[sigplan,10pt,nonacm]{acmart}

\AtBeginDocument{%
  }

\usepackage{algorithm}
\usepackage{algorithmic}
\usepackage{subcaption}
\usepackage{enumitem}

\newcommand{\SYSNAME}{\textsc{StreamKL}}

\setcopyright{acmlicensed}
\copyrightyear{2018}
\acmYear{2018}
\acmDOI{XXXXXXX.XXXXXXX}
\acmConference[Conference acronym 'XX]{Make sure to enter the correct
  conference title from your rights confirmation email}{June 03--05,
  2018}{Woodstock, NY}
\acmISBN{978-1-4503-XXXX-X/2018/06}

\title[\SYSNAME{}]{\SYSNAME{}: Fast and Memory-Efficient KL Divergence for Boosting Attention Distillation}

\author{Guangda Liu}
\affiliation{\institution{Shanghai Jiao Tong University}\country{}}

\author{Yiquan Wang}
\affiliation{\institution{Shanghai Jiao Tong University}\country{}}

\author{Chengwei Li}
\affiliation{\institution{Shanghai Jiao Tong University}\country{}}

\author{Wenhao Chen}
\affiliation{\institution{Shanghai Jiao Tong University}\country{}}

\author{Jing Lin}
\affiliation{\institution{Huawei}\country{}}

\author{Yiwu Yao}
\affiliation{\institution{Huawei}\country{}}

\author{Danning Ke}
\affiliation{\institution{Huawei}\country{}}

\author{Wenchao Ding}
\affiliation{\institution{Fudan University}\country{}}

\author{Jieru Zhao}
\authornote{Corresponding author: zhao-jieru@sjtu.edu.cn.}
\affiliation{\institution{Shanghai Jiao Tong University}\country{}}

\renewcommand{\shortauthors}{Liu et al.}

\begin{abstract}
Attention distillation, which trains one attention distribution to match another by minimizing their Kullback-Leibler (KL) divergence, is widely used in knowledge distillation, model compression, continual learning, and sparse-attention LLM training. 
However, existing approaches materialize both attention distributions before computing the KL reduction, incurring $O(N_QN_K)$ memory and IO costs that become prohibitive at long context lengths. 
We present \SYSNAME{}, the first fused GPU primitive for attention KL divergence that eliminates this quadratic materialization.
\SYSNAME{} derives a novel online formulation for the coupled two-distribution KL reduction, enabling a single one-pass forward kernel that streams query-key tiles through on-chip SRAM.
For the backward pass, \SYSNAME{} recomputes attention probabilities tile-by-tile, avoiding storage of quadratic intermediates.
We further design and implement efficient GPU kernels with dedicated optimizations.
Experiments show \SYSNAME{} delivers up to $43\times$ and $14\times$ speedups over baseline methods in the forward and backward passes, respectively.
Most importantly, \SYSNAME{} reduces the extra HBM footprint of attention distillation from $O(N_QN_K)$ to $O(1)$, enabling long-context distillation on a single GPU.
\end{abstract}

\begin{document}
\maketitle

\section{Introduction}

Attention distillation, which trains one attention distribution to match another by minimizing their Kullback-Leibler (KL) divergence, is a core building block of modern transformer workloads.
In knowledge distillation, it transfers attention patterns from a large teacher model to a smaller student model~\cite{minilm}.
In model pruning, it aligns the attention distributions of the pruned model with those of the dense model to preserve accuracy~\cite{D2Prune}.
In continual learning, it aligns the attention distributions of the current model with a historical snapshot to mitigate catastrophic forgetting~\cite{agrawal-etal-2025-multilingual}.
Most recently, during the training of sparse-attention LLMs such as DeepSeek V3.2 and GLM-5, attention distillation is used to align a lightweight indexer distribution with the original dense-attention distribution~\cite{deepseek-v32,glm-5}.
Across all of these settings, the training objective is to minimize the KL divergence between two attention distributions $P_1, P_2 \in \mathbb{R}^{N_Q \times N_K}$ generated from respective query-key pairs $Q_1 \in \mathbb{R}^{N_Q \times d_1}, K_1 \in \mathbb{R}^{N_K \times d_1}$ and $Q_2 \in \mathbb{R}^{N_Q \times d_2}, K_2 \in \mathbb{R}^{N_K \times d_2}$.

\begin{figure}[t]
  \centering
  \includegraphics[width=\linewidth]{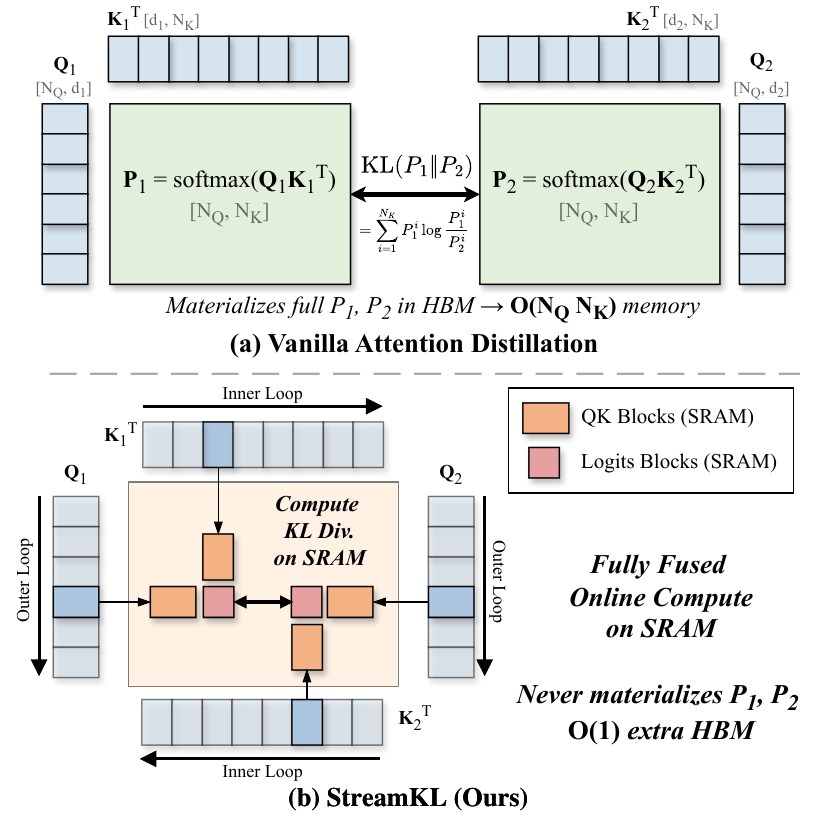}
  \caption{Overview of \SYSNAME{}. (a) Vanilla attention distillation materializes full $P_1, P_2$ in HBM, costing $O(N_Q N_K)$ memory and IO. (b) \SYSNAME{} fuses the computation into a one-pass tiled kernel that computes KL online in SRAM without materializing $P_1$ or $P_2$, reducing extra HBM to $O(1)$.}
  \label{fig:intro}
\end{figure}

A na\"ive approach to attention distillation first materializes both attention distributions in HBM and then computes the element-wise KL reduction, as shown in Fig.~\ref{fig:intro}(a).
This works at short context lengths; however, as the context grows, it faces significant performance challenges.
First, the \textit{memory footprint} explodes: at $N_Q = N_K = 64\text{K}$ with 32 attention heads in BF16, storing the two distributions alone requires \textbf{512\,GB}, which is \textbf{3.6 times} the 141\,GB HBM capacity of a single NVIDIA H200 GPU.
Second, the \textit{IO traffic} of writing these distributions out and reading them back for the KL reduction dominates runtime.
Both of the costs scale \textbf{quadratically} with the context length.
As modern LLMs push context lengths toward 128K, 256K, and even beyond 1M tokens~\cite{deepseek-v32,glm-5}, long-context attention distillation becomes a severe system bottleneck.
A common workaround is chunked processing, which iterates over query chunks sequentially so that only a slice of $P_1$ and $P_2$ is materialized at a time.
However, this trades latency for memory with no good operating point: small chunks leave the GPU idle, while large chunks trigger out-of-memory (OOM) errors.
Fundamentally, \textbf{these bottlenecks cannot be resolved without eliminating the $O(N_Q N_K)$ materialization of $P_1$ and $P_2$}. 

Standard attention computation once faced a similar quadratic memory and IO problem, which FlashAttention~\cite{flash-attention,flash-attn2} solved by fusing the entire computation into a single \textit{online-update} kernel that never materializes the $N_Q \times N_K$ matrix.
One might hope to apply a similar recipe to attention distillation, however, the adaptation is non-trivial.
Directly leveraging online softmax either still materializes $O(N_Q N_K)$ intermediates, or requires two separate passes, doubling both computation and HBM reads.
The key difficulty in achieving one-pass fusion is that, unlike standard attention which applies a single softmax followed by a matrix multiply, KL divergence couples \emph{two} independent softmax distributions through a weighted logit-difference term, and how to correctly rescale the accumulated logit-difference as the running maxima of both distributions change is not straightforward to derive.
To the best of our knowledge, \textbf{no existing system provides a fused primitive for attention distillation that avoids materializing $P_1$ and $P_2$.}

To eliminate the quadratic memory and IO costs of attention distillation at the root, we present \SYSNAME{}, the first fused, one-pass primitive for attention KL divergence.
For the forward pass, \SYSNAME{} derives a new online-update scheme for the coupled two-distribution KL reduction.
Through a careful mathematical derivation, we reformulate the row-wise KL divergence into a form that can be computed incrementally in a single streaming pass over the keys.
For the backward pass, \SYSNAME{} uses \textit{recomputation} over the saved log-sum-exp (LSE) vectors to regenerate attention probabilities tile-by-tile, without storing the quadratic intermediates.
As illustrated in Fig.~\ref{fig:intro}, with complete fused online computation on SRAM, \SYSNAME{} reduces the extra HBM footprint of attention distillation from $O(N_Q N_K)$ to $O(1)$.

Beyond the algorithmic contribution, \SYSNAME{} designs and implements efficient GPU kernels, incorporating the optimizations for small-query workloads, such as the auto-regressive decoding in LLMs where $N_Q=1$.
For the forward pass, a \textit{split-K} variant partitions the key dimension across thread blocks to saturate SMs.
For the backward pass, a \textit{fused} kernel eliminates redundant HBM traffic and logit recomputation by visiting each $QK$ tile pair exactly once.
On Hopper GPUs, dedicated kernels further leverage the Tensor Memory Accelerator (TMA) for asynchronous bulk copies.

We evaluate \SYSNAME{} on NVIDIA H200 and A100 GPUs across various workloads.
In the forward pass, \SYSNAME{} achieves up to $18\times$ speedup over PyTorch, $3.7\times$ over \texttt{torch.} \texttt{compile}, and $3.7\times$ over FLA, widening to $43\times$, $7.0\times$, and $7.1\times$ under causal masking.
The backward pass delivers up to $6.5\times$ over PyTorch on non-causal inputs and $14.0\times$ under causal masking.
Most importantly, \SYSNAME{} reduces the extra HBM footprint from $O(N_Q N_K)$ to $O(1)$, making it the only solution that sustains 64K+ contexts on a single GPU.

In summary, we make the following contributions:
\begin{itemize}
  \item We identify the quadratic memory and IO bottlenecks that prevent attention KL divergence from scaling to long contexts (Sec.~\ref{sec:moti}).
  \item We introduce the first online reformulation of attention KL divergence, enabling a fully fused online-update forward primitive (Sec.~\ref{sec:fwd}), together with a recomputation-based backward pass. That eliminates the materialization of quadratic intermediates (Sec.~\ref{sec:bwd}).
  \item We design and implement efficient GPU kernels for \SYSNAME{} with dedicated optimizations (Sec.~\ref{sec:impl}).
   Extensive experiments show that \SYSNAME{} achieves order-of-mag\-nitude improvements in both latency and peak HBM footprint, enabling previously infeasible long-context attention-distillation workloads (Sec.~\ref{sec:eval}).
\end{itemize}

\section{Background}

\subsection{Attention Distillation} \label{sec:bg-ad}
Scaled dot-product attention (SDPA) is a core component of modern machine learning models, including LLMs, DiTs, ViTs, and other Transformer-based architectures~\cite{transformer,gpt,dit,vit}.
Attention distillation is a common technique in which one attention distribution is trained to match another.
Given two query-key pairs, $Q_1\in \mathbb{R}^{N_Q \times d_1}, K_1\in \mathbb{R}^{N_K \times d_1}$ and $Q_2\in \mathbb{R}^{N_Q \times d_2}, K_2\in \mathbb{R}^{N_K \times d_2}$, let $P_1 = \text{softmax}(Q_1 K_1^T)$ and $P_2 = \text{softmax}(Q_2 K_2^T)$ denote the attention distributions ($\sqrt{d}$ omitted for brevity), where $P_1, P_2\in \mathbb{R}^{N_Q \times N_K}$.
Attention distillation then minimizes the KL divergence between these distributions, i.e., $\mathrm{KL}(P_1 \| P_2)$, as the training objective.

Attention distillation arises in a wide range of scenarios.
In knowledge distillation, MiniLM~\cite{minilm} transfers attention patterns from teacher to student by distilling SDPA over query-key and value-value interactions.
In model compression, attention distillation guides pruned or quantized models to preserve the original model's attention distributions~\cite{D2Prune,miniq}.
Beyond LLMs, DiT models exploit attention KL divergence for unsupervised segmentation, personalized generation, and reducing diffusion timesteps~\cite{jo2026tracediffusionmodelsecretly,tian2024diffuseattendsegmentunsupervised,lim2025conceptsplitdecoupledmulticonceptpersonalization}.

Another important application of attention distillation is training sparse-attention LLMs.
For example, DeepSeek Sparse Attention (DSA), used in DeepSeek V3.2 and GLM-5, employs a lightning indexer to estimate token importance for sparse attention computation~\cite{deepseek-v32,glm-5}.
During training, attention distillation is used to align the lightning indexer distribution with the original dense-attention distribution.

\paragraph{KL Divergence}
In batched multi-head attention, the training objective is the mean KL divergence over batch, head, and query rows. 
For clarity, we present the formulation for a single query row with $q_1\in\mathbb{R}^{1\times d_1}$ and $q_2\in\mathbb{R}^{1\times d_2}$.
Let the logits be $S_1=q_1K_1^T\in\mathbb{R}^{1\times N_K}$ and $S_2=q_2K_2^T\in\mathbb{R}^{1\times N_K}$, and define the corresponding attention distributions as $P_1=\mathrm{softmax}(S_1)\in\mathbb{R}^{1\times N_K}$ and $P_2=\mathrm{softmax}(S_2)\in\mathbb{R}^{1\times N_K}$.
The forward computation of the row-wise KL divergence is 
\begin{equation}
\label{eq:kl-def}
L=\mathrm{KL}(P_1\|P_2)=\sum_{i=1}^{N_K}P_1^i \log \frac{P_1^i}{P_2^i}=\sum_{i=1}^{N_K} P_1^i\big(\log P_1^i-\log P_2^i\big).
\end{equation}
Here, $P_1^i$ and $P_2^i$ denote the $i$-th token probabilities of $P_1$ and $P_2$, respectively.
During backpropagation, we consider two settings depending on which distribution is treated as fixed.
In Setting 1, $P_1$ is fixed and $P_2$ is optimized (standard distillation). The gradients are computed for $q_2$ and $K_2$:
\begin{equation}
\label{eq:kl-case1}
dS_2 = \frac{\partial L}{\partial S_2} = P_2 - P_1,\quad
dq_2 = dS_2K_2,\quad
dK_2 = dS_2^Tq_2.
\end{equation}
In Setting 2, $P_2$ is fixed and $P_1$ is optimized. To compute gradients for $q_1$ and $K_1$, we define an intermediate variable $r=\log P_1-\log P_2\in\mathbb{R}^{1\times N_K}$, and obtain
\begin{equation}
\label{eq:kl-case2}
dS_1 = \frac{\partial L}{\partial S_1}=P_1\odot(r-L),\ dq_1 = dS_1K_1,\ dK_1 = dS_1^Tq_1.
\end{equation}

\subsection{Online Softmax and FlashAttention}

\paragraph{Online Softmax.}
The softmax function $P^i = \frac{\exp(S^i)}{\sum_j \exp(S^j)}$ poses a fundamental challenge for single-pass, tiled computation: the normalization constant $\sum_j \exp(S^j)$ depends on all $N_K$ elements, so a na\"ive implementation requires two sequential passes over the input to compute the maximum and denominator, and another to normalize.
Milakov and Gimelshein~\cite{online-softmax} introduced the \emph{online softmax} algorithm, which reduces this to a single pass by maintaining two running statistics: the current maximum $m$ and the unnormalized sum of exponentials $l$.
Concretely, for each new tile of logits $S'$, the statistics are updated as follows:
\begin{equation}
\label{eq:online-softmax}
m' = \max(m,\, \max(S')), \quad
l' = l \cdot e^{m - m'} + \sum_{j \in \text{tile}} e^{S^j - m'}.
\end{equation}
When a new tile reveals a larger maximum $m'$, the accumulated sum $l$ is rescaled by the correction factor $e^{m - m'}$ to maintain a consistent reference point.
After all tiles have been processed, the log-sum-exp is recovered as $\mathrm{LSE} = m + \log l$, and the probability can be obtained as $P^i = \exp(S^i - \mathrm{LSE})$ without storing the full probabilities.
This online formulation enables computing softmax in a tiled, streaming fashion.

\paragraph{FlashAttention.}
Modern GPUs feature a memory hierarchy comprising a small but fast on-chip SRAM and a large but slow off-chip HBM.
Standard attention computes the $N_Q \times N_K$ attention matrix $S = QK^T$, writes $P = \mathrm{softmax}(S)$ to HBM, and then reads it back to compute the output $O = PV$.
This requires $O(N_Q N_K)$ HBM storage and memory traffic, which dominates both the runtime and the memory footprint for long sequences.
FlashAttention~\cite{flash-attention,flash-attn2} eliminates this bottleneck by fusing the entire attention computation into a single kernel that never materializes $P$ in HBM.
The algorithm partitions $Q$ and $K$ into tiles and processes the tiles in a streaming manner.
For each $QK$ tile, the kernel computes a block of logits entirely in on-chip SRAM, applies the online softmax update to maintain running statistics $(m, l)$, and immediately accumulates the partial output by rescaling the accumulated output.
Once all tiles have been processed, the final output is written to HBM.
This tiling strategy largely reduces the HBM access complexity, while simultaneously reducing the memory footprint from $O(N_Q N_K)$ to $O(N_Q)$.

The backward pass of FlashAttention applies the same tiling philosophy as online softmax.
Rather than storing $P$ from the forward pass, it saves only the per-row LSE values and recomputes $P$ block-by-block from $Q$ and $K$ during backpropagation.
This recomputation-over-storage trade-off is favorable because the recomputation cost is dominated by fast on-chip matrix multiplies, whereas reading a stored $O(N_Q N_K)$ matrix from HBM would be far more expensive.

\subsection{Motivation} \label{sec:moti}

Similar to attention computation, applying attention distillation with long contexts presents severe system challenges.
A na\"ive implementation first materializes both attention distributions $P_1, P_2 \in \mathbb{R}^{N_Q \times N_K}$ in HBM, then computes the element-wise KL divergence.
This incurs the following two costs that scale quadratically with the context length.

\begin{figure}[t]
  \centering
  \includegraphics[width=\linewidth]{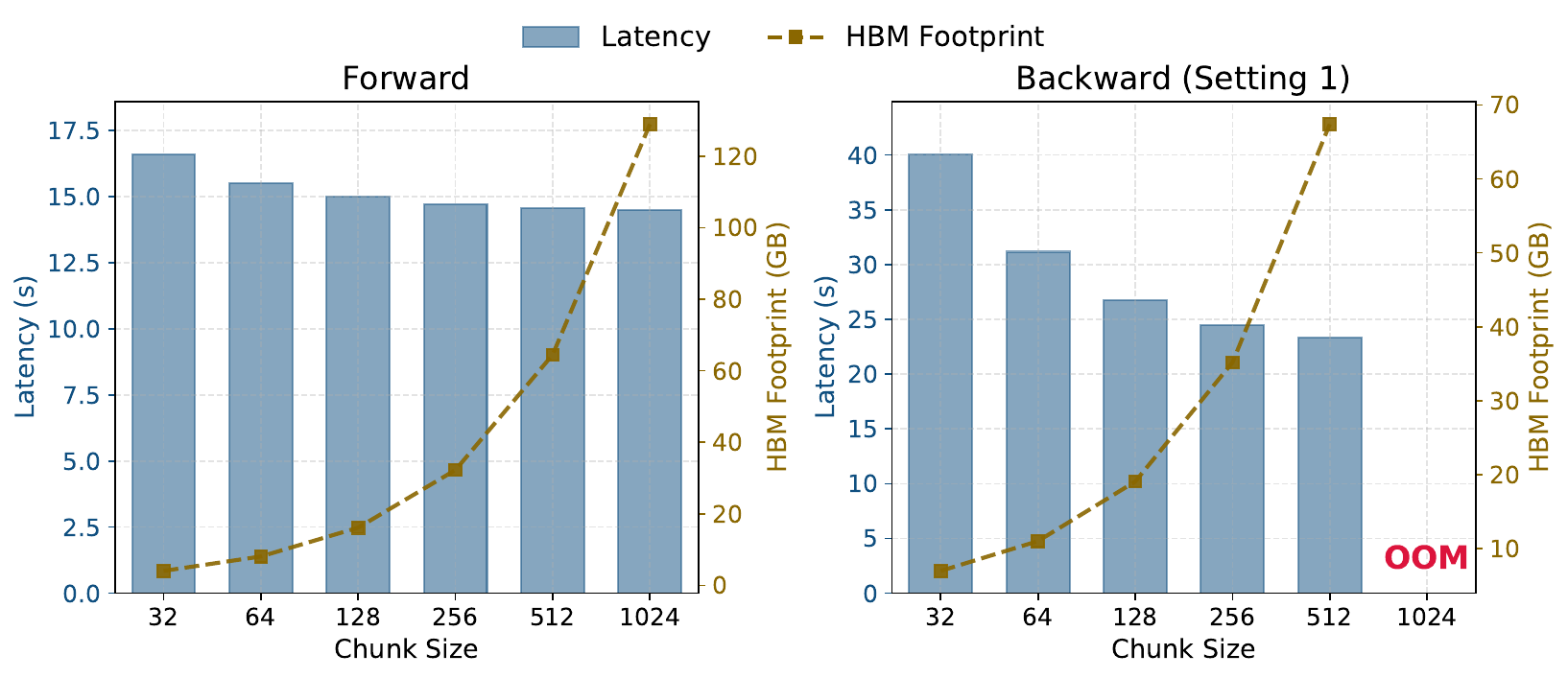}
  \caption{Latency-memory trade-off of chunked attention KL divergence, with $N_Q = N_K = 128\text{K}$ and batch size 32.}
  \label{fig:chunk-tradeoff}
\end{figure}

\paragraph{Memory Footprint.}
Materializing both $P_1$ and $P_2$ requires $2 \times N_Q \times N_K$ elements in HBM, scaling quadratically with the context length.
For a typical long-context setup with $N_Q = N_K = 64\text{K}$, 32 attention heads, and BF16 precision, the two distributions alone consume $2 \times 32 \times 64\text{K} \times 64\text{K} \times 2\text{B} = \textbf{512\,GB}$ of HBM, which is \textbf{3.6 times} the capacity of a single H200 (141\,GB).
In sparse-attention LLM training, context lengths can reach 128K, 256K, or even 1M~\cite{deepseek-v32,glm-5}, further amplifying the footprint.
Even at shorter contexts where both distributions fit, the quadratic footprint crowds out activations and optimizer states, forcing smaller batch sizes or sharding across devices, both of which hurt training throughput.
To reduce the memory footprint, a straightforward workaround is chunked processing: iterating over query chunks sequentially so that only slices of $P_1, P_2$ are materialized at a time.
However, chunked processing faces an intrinsic dilemma between latency and memory.
Figure~\ref{fig:chunk-tradeoff} illustrates this trade-off tested on an H200 with batch size 32 and $N_Q = N_K = 128\text{K}$.
Smaller chunks reduce peak HBM footprint but increase latency, as fewer query rows per launch limit GPU parallelism.
Larger chunks improve throughput but the memory footprint grows rapidly.
At chunk size 512, the backward pass already consumes about 70\,GB, and at chunk size 1024 it triggers OOM on the 141\,GB H200.

\begin{figure}[t]
  \centering
  \includegraphics[width=\linewidth]{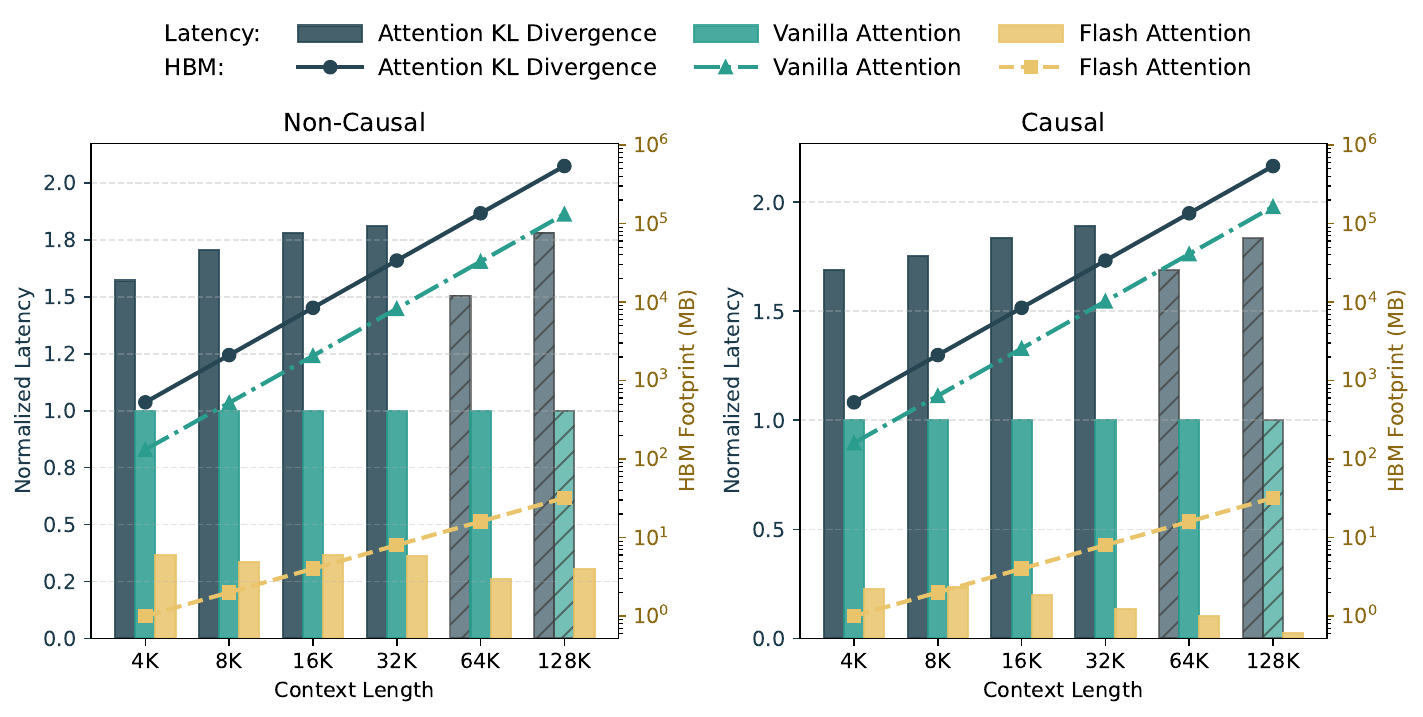}
  \caption{Normalized latency and HBM footprint of attention KL divergence, vanilla attention, and FlashAttention (H200, batch 32). Hatched bars denote OOM with extrapolation.}
  \label{fig:io-bottleneck}
\end{figure}

\paragraph{IO Bottleneck.}
Even when both $P_1$ and $P_2$ fit in HBM, IO traffic can dominate the runtime of attention distillation.
The forward pass writes $2 \times N_Q \times N_K$ values to HBM and reads them back for the KL reduction.
Figure~\ref{fig:io-bottleneck} compares the latency and HBM footprint of attention KL divergence against vanilla attention and FlashAttention across context lengths from 4K to 128K.
Since KL divergence materializes two $N_Q \times N_K$ matrices, its IO volume and latency are consistently $1.5$--$2\times$ that of vanilla attention.
FlashAttention, by contrast, eliminates the $N_Q \times N_K$ materialization entirely via a fused kernel, achieving significantly lower latency and HBM footprint than vanilla attention.
The advantage is even more pronounced under causal masking, where FlashAttention skips masked positions while the vanilla implementations still materialize and mask the full $N_Q \times N_K$ matrix.

FlashAttention demonstrates that fusing the attention computation into a single one-pass kernel and eliminating the $N_Q \times N_K$ materialization can yield order-of-magnitude improvements in both latency and memory.
This naturally raises the question: \emph{can a similar IO-aware fusion be applied to attention KL divergence?}
The answer is non-trivial.
Directly leveraging online softmax either still materializes $O(N_Q N_K)$ intermediates for the logit matrix $S$, or requires two separate passes (one to obtain the LSE values and another to compute the KL reduction), doubling both computation and HBM reads.
For a one-pass fusion, the key difficulty is that, unlike standard attention which applies a single softmax followed by a matrix multiply, KL divergence couples \emph{two} independent softmax distributions through a weighted logit-difference term.
How to correctly rescale the accumulated weighted logit-difference as the running maxima of both distributions change is not straightforward to derive.
\textbf{No existing system provides such a fused attention-KL primitive.}
\SYSNAME{} addresses this challenge with a novel reformulation of attention KL divergence that enables one-pass, online computation over both distributions in a single fused kernel, eliminating all quadratic materializations and reducing the extra HBM footprint from $O(N_Q N_K)$ to $O(1)$.

\section{Forward Pass}
\label{sec:fwd}

\subsection{Formulation for Online Computation}

To evaluate $\mathrm{KL}(P_1 \| P_2)$ without materializing the $N_Q \times N_K$ attention distributions, we reformulate the KL divergence in online-computable form.
Recall from Eq.~\eqref{eq:kl-def}, the row-wise KL divergence is $L = \sum_{i=1}^{N_K} P_1^i \log(P_1^i / P_2^i)$.
Using the logits $S_1 = q_1 K_1^T$ and $S_2 = q_2 K_2^T$, we write the softmax probabilities in terms of the online-softmax running statistics: the row-wise maximum $m_1 = \max_j S_1^j$ and $m_2 = \max_j S_2^j$, and the unnormalized sums $l_1 = \sum_j e^{S_1^j - m_1}$ and $l_2 = \sum_j e^{S_2^j - m_2}$:
\begin{equation}
\label{eq:prob-from-stats}
P_1^i = \frac{e^{S_1^i - m_1}}{l_1}, \quad P_2^i = \frac{e^{S_2^i - m_2}}{l_2}.
\end{equation}

Substituting Eq.~\eqref{eq:prob-from-stats} into the KL divergence definition and expanding the log-ratio yields:
\begin{align}
L &= \sum_i \frac{e^{S_1^i - m_1}}{l_1} \log \frac{e^{S_1^i - m_1} / l_1}{e^{S_2^i - m_2} / l_2} \nonumber \\
  &= \sum_i \frac{e^{S_1^i - m_1}}{l_1} \Big[\underbrace{\log e^{S_1^i - m_1} - \log e^{S_2^i - m_2}}_{= \;(S_1^i - m_1) - (S_2^i - m_2)} + \underbrace{\log \frac{l_2}{l_1}}_{\text{constant w.r.t.\ } i}\Big] \nonumber \\
  &= \sum_i \frac{e^{S_1^i - m_1}}{l_1} \big(S_1^i - S_2^i - m_1 + m_2\big) + \log \frac{l_2}{l_1}. \label{eq:kl-expanded}
\end{align}
In the last step, we used $\sum_i \frac{e^{S_1^i - m_1}}{l_1} = \sum_i P_1^i = 1$ to factor out the $\log \frac{l_2}{l_1}$ term.
Similarly, we now separate the constant $(m_2 - m_1)$ from the $i$-dependent logit difference by splitting the sum in Eq.~\eqref{eq:kl-expanded}:
\begin{align}
L &= \frac{1}{l_1} \sum_i e^{S_1^i - m_1} (S_1^i - S_2^i) + \underbrace{(m_2 - m_1) \sum_i \frac{e^{S_1^i - m_1}}{l_1}}_{= \; m_2 - m_1} + \log \frac{l_2}{l_1} \nonumber \\
  &= \frac{1}{l_1}\underbrace{\sum_i e^{S_1^i - m_1}(S_1^i - S_2^i)}_{\displaystyle \triangleq\;\mathrm{acc}} + (m_2 - m_1) + \log l_2 - \log l_1. \label{eq:kl-acc}
\end{align}
Here we define the \emph{accumulator} $\mathrm{acc} = \sum_i e^{S_1^i - m_1}(S_1^i - S_2^i)$, which captures the weighted logit difference and must be maintained during the tiled pass.

Finally, recalling that $\mathrm{LSE}_t = m_t + \log l_t$, we combine $(m_2 - m_1) + \log(l_2/l_1) = \mathrm{LSE}_2 - \mathrm{LSE}_1$ and arrive at our online-computable objective:
\begin{equation}
\label{eq:kl-final}
\boxed{\;L = \mathrm{KL}(P_1 \| P_2) = \frac{\mathrm{acc}}{l_1} + \mathrm{LSE}_2 - \mathrm{LSE}_1.\;}
\end{equation}
This formulation is the foundation of the \SYSNAME{} forward pass: the five scalars $(m_1, l_1, m_2, l_2, \mathrm{acc})$ can all be online updated as each $K$ tile is streamed through SRAM, using a similar online rescaling mechanism as Eq.~\eqref{eq:online-softmax}.
The detailed algorithm design is presented in Section~\ref{sec:fwd-algorithm}.

\paragraph{Generalizability beyond Softmax}
\SYSNAME{} readily extends to attention activation functions beyond softmax, such as the ReLU variant used in the indexer training of DeepSeek-V3.2~\cite{deepseek-v32}.
In fact, activation functions that lack softmax's global normalization, such as ReLU, sigmoid, or tanh, are even simpler to support, as they remove the need for maintaining multiple global normalizations.

\subsection{Algorithm Design} \label{sec:fwd-algorithm}

Algorithm~\ref{alg:flashkl-fwd} presents the \SYSNAME{} forward pass.
Thanks to the reformulation in Eq.~\ref{eq:kl-acc}, the coupled two-distribution KL reduction is expressed in terms of five per-row running scalars $(m_1, l_1, m_2, l_2, \mathrm{acc})$ that can be updated incrementally as key tiles are streamed through SRAM, with joint rescaling whenever either running maximum changes.
The outer loop partitions query rows across thread blocks; the inner loop co-streams $K_1$ and $K_2$ tiles sequentially. 

\begin{algorithm}[t]
  \caption{\small \SYSNAME{} Forward Pass \label{alg:flashkl-fwd}}
  \begin{algorithmic}[1]
    \small
    \REQUIRE $Q_1 \in \mathbb{R}^{N_Q \times d_1}$, $K_1 \in \mathbb{R}^{N_K \times d_1}$, $Q_2 \in \mathbb{R}^{N_Q \times d_2}$, $K_2 \in \mathbb{R}^{N_K \times d_2}$ in HBM, block sizes $B_Q$, $B_K$.
    \STATE Divide $Q_1, Q_2$ into $T_Q = \lceil N_Q / B_Q \rceil$ blocks of size $B_Q$; divide $K_1, K_2$ into $T_K = \lceil N_K / B_K \rceil$ blocks of size $B_K$.
    \FOR{$1 \le i \le T_Q$ \textbf{in parallel}}
      \STATE Load $Q_{1,i}, Q_{2,i}$ from HBM to SRAM.
      \STATE Initialize on chip:
        $m_1^{(0)}, m_2^{(0)} \leftarrow (-\infty)_{B_Q}$;\;
        $l_1^{(0)}, l_2^{(0)} \leftarrow (0)_{B_Q}$;\;
        $\mathrm{acc}^{(0)} \leftarrow (0)_{B_Q}$.
      \FOR{$1 \le j \le T_K$}
        \STATE Load $K_{1,j}, K_{2,j}$ from HBM to SRAM.
        \STATE Compute logit tiles $S_1 = Q_{1,i}\, K_{1,j}^T$, $S_2 = Q_{2,i}\, K_{2,j}^T$.
        \STATE Update maximums:\\
        $m_1^{(j)} = \max(m_1^{(j-1)},\, \mathrm{rowmax}(S_1))$,\\
        $m_2^{(j)} = \max(m_2^{(j-1)},\, \mathrm{rowmax}(S_2))$.
        \STATE Compute correction factors:\\
        $\alpha_1 = e^{m_1^{(j-1)} - m_1^{(j)}}$,
        $\alpha_2 = e^{m_2^{(j-1)} - m_2^{(j)}}$.
        \STATE Compute unnormalized probabilities:\\
        $\tilde{P}_1 = e^{S_1 - m_1^{(j)}} \in \mathbb{R}^{B_Q \times B_K}$,
        $\tilde{P}_2 = e^{S_2 - m_2^{(j)}} \in \mathbb{R}^{B_Q \times B_K}$.
        \STATE Update running sums:\\
        $l_1^{(j)} = \alpha_1 \, l_1^{(j-1)} + \mathrm{rowsum}(\tilde{P}_1)$,\\
        $l_2^{(j)} = \alpha_2 \, l_2^{(j-1)} + \mathrm{rowsum}(\tilde{P}_2)$.
        \STATE Update accumulator:\\
        $\mathrm{acc}^{(j)} = \alpha_1 \,\mathrm{acc}^{(j-1)} + \mathrm{rowsum}\big(\tilde{P}_1 \odot (S_1 - S_2)\big)$.
      \ENDFOR
      \STATE Compute $\mathrm{LSE}_1 = m_1^{(T_K)} + \log l_1^{(T_K)}$,\;\\ 
      Compute $\mathrm{LSE}_2 = m_2^{(T_K)} + \log l_2^{(T_K)}$.
      \STATE Compute $\mathrm{KL}_i = \mathrm{acc}^{(T_K)} / l_1^{(T_K)} + \mathrm{LSE}_2 - \mathrm{LSE}_1$.
      \STATE Write $\mathrm{KL}_i$, $\mathrm{LSE}_{1}$, $\mathrm{LSE}_{2}$ to HBM.
    \ENDFOR
    \RETURN KL values $\in \mathbb{R}^{N_Q}$ and saved $\mathrm{LSE}_1, \mathrm{LSE}_2 \in \mathbb{R}^{N_Q}$.
  \end{algorithmic}
\end{algorithm}

\paragraph{Accumulator rescaling invariant.}
We establish correctness by verifying the central loop invariant.
After processing key tiles $1, \dots, j$, the accumulator satisfies:
\begin{equation}
\label{eq:acc-invariant}
\mathrm{acc}^{(j)} = \sum_{i \in \text{tiles } 1..j} e^{S_1^i - m_1^{(j)}} (S_1^i - S_2^i).
\end{equation}
where $m_1^{(j)}$ is the running row-wise logit maximum of $P_1$ and $\mathrm{acc}^{(j)}$ is the unnormalized logit-difference sum, both over tiles $1..j$.
When tile $j{+}1$ raises the maximum, the accumulator is rescaled by the correction factor $\alpha_1 = e^{m_1^{(j)} - m_1^{(j+1)}}$:
\begin{align}
\mathrm{acc}^{(j+1)}
&= \alpha_1 \,\mathrm{acc}^{(j)} + \sum_{i \in \text{tile } j+1} e^{S_1^i - m_1^{(j+1)}} (S_1^i - S_2^i) \nonumber \\
&= \sum_{i \in \text{tiles } 1..j+1} e^{S_1^i - m_1^{(j+1)}} (S_1^i - S_2^i). \label{eq:acc-update}
\end{align}
By induction, at the end of the inner loop ($j = T_K$), $\mathrm{acc}^{(T_K)}$ matches $\sum_{i=1}^{N_K} e^{S_1^i - m_1} (S_1^i - S_2^i)$ as required by Eq.~\eqref{eq:kl-acc}.
The running sums $l_1$ and $l_2$ follow standard online softmax rescaling (Eq.~\ref{eq:online-softmax}) using $\alpha_1$ and $\alpha_2$, respectively.

Note that unlike FlashAttention, which must rescale a $B_Q \times d$ output matrix during each running maximum update, our accumulator $\mathrm{acc}$ is a $B_Q$-dimensional vector. 
This reduces the arithmetic rescaling overhead by a factor of $d$ (typically at least 128).
As recent studies such as FlashAttention-4~\cite{flashattention4} indicate that rescaling arithmetic can become a prominent bottleneck on frontier hardware such as NVIDIA Blackwell, \SYSNAME{}'s vector-only rescaling ensures that the kernel can fully utilize tensor core throughput without being constrained by scalar functional units.

\paragraph{Causal masking.}
When causal masking is applied to the attention distributions, \SYSNAME{} exploits the triangular attention structure to avoid unnecessary computation.
Costly element-wise masking operations are strictly isolated to the partially masked boundary logit tiles along the diagonal.
The inner loop terminates early to skip all key tiles located entirely beyond the causal frontier.

\paragraph{Saved state for backward.}
To avoid materializing the full $O(N_Q N_K)$ attention matrices, \SYSNAME{} relies on recomputation.
The forward pass writes only the final KL values and the log-sum-exp vectors $\mathrm{LSE}_1$, $\mathrm{LSE}_2$ to HBM, storing $O(N_Q)$ scalars.
The backward pass regenerates $P_1^i = e^{S_1^i - \mathrm{LSE}_1}$ and $P_2^i = e^{S_2^i - \mathrm{LSE}_2}$ on the fly from the original $Q$, $K$ and saved LSEs, reducing activation memory from $O(N_Q N_K)$ to $O(1)$.

\paragraph{IO complexity.}
Each query tile loads $Q_{1,i} \in \mathbb{R}^{B_Q \times d_1}$, $Q_{2,i} \in \mathbb{R}^{B_Q \times d_2}$ once and streams all $T_K$ key tiles, giving total HBM reads $O(N_Q N_K(d_1+d_2)/B_Q)$. 
The HBM writes only consist of $3N_Q$ scalars (KL, $\mathrm{LSE}_1$, $\mathrm{LSE}_2$).

\subsection{GPU Kernel Design}  \label{sec:fwd-kernel}

\begin{figure}[t]
  \centering
  \includegraphics[width=\columnwidth]{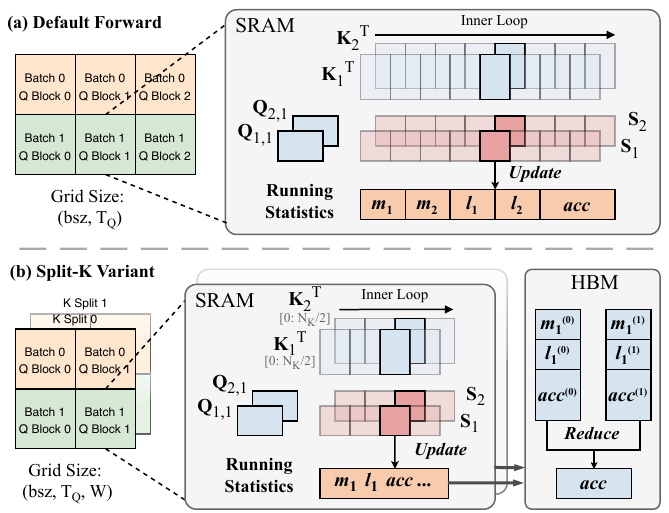}
  \caption{Forward kernel design. (a) Default kernel on a $(\mathrm{bsz}, T_Q)$ grid. (b) Split-K variant: a third grid dimension $W$ partitions the $K$ dimension; each block writes its partial statistics to HBM and a lightweight reduce merges them.}
  \label{fig:fwd-overview}
\end{figure}

\paragraph{Default kernel.}
The default forward kernel directly instantiates Algorithm~\ref{alg:flashkl-fwd}, where each thread block processes one query tile and streams all key tiles sequentially, as illustrated in Figure~\ref{fig:fwd-overview}(a).
The grid is two-dimensional, indexed by $(b, i)$ where $b$ is the batch/head index and $i$ is the query-tile index, yielding $\mathrm{bsz} \times T_Q$ thread blocks in total.
The default kernel is efficient when $\mathrm{bsz} \times T_Q$ is large enough to saturate the GPU's streaming multiprocessors (SMs).
However, when $\mathrm{bsz} \times T_Q$ falls well below the SM count, the GPU is underutilized.
For instance, with $\mathrm{bsz}=1$, $N_Q=512$, and $B_Q=32$, only 16 thread blocks are launched on an H200 with 132 SMs---leaving over 85\% of the SMs idle.
The resulting low occupancy leaves the GPU unable to hide memory latency, severely underutilizing compute throughput.

\paragraph{Split-K variant.}
To recover full GPU utilization in these low-parallelism regimes, we introduce a \emph{split-K} variant that creates additional parallelism along the key dimension, as illustrated in Figure~\ref{fig:fwd-overview}(b).
The key range $[1, N_K]$ is divided into $W$ contiguous chunks of size $C_K = \lceil N_K / W \rceil$, and a third grid dimension is added so that the launch grid becomes $(b, i, w)$ with $\mathrm{bsz} \times T_Q \times W$ thread blocks.
Each partial thread block $(b, i, w)$ processes only its assigned key chunk and writes its local running statistics $(m_1, l_1, \mathrm{acc})$ to a temporary buffer in HBM.

A lightweight \emph{reduce} kernel then merges the $W$ partial results for each $(b, i)$ pair.
The merge applies a similar rescaling used in the inner loop: given two partial results $a$ and $b$, the combined statistics are:
\begin{equation}
m_1' = \max(m_1^{(a)}, m_1^{(b)}), \qquad
l_1' = l_1^{(a)} e^{m_1^{(a)} - m_1'} + l_1^{(b)} e^{m_1^{(b)} - m_1'} \nonumber
\end{equation}
\begin{equation}
\mathrm{acc}' = \mathrm{acc}^{(a)} e^{m_1^{(a)} - m_1'} + \mathrm{acc}^{(b)} e^{m_1^{(b)} - m_1'} 
\end{equation}
Since the accumulator rescaling (Eq.~\ref{eq:acc-invariant}) is associative, the merged result is identical to the original forward kernel.

\paragraph{Adaptive split selection.}
The split count $W$ is determined at launch time to reach a target occupancy of $P_{\mathrm{target}}$ thread blocks (typically 128 or 256).
Let $P_{\mathrm{base}} = \mathrm{bsz} \times T_Q$ denote the baseline thread-block count of the non-split kernel.
When $P_{\mathrm{base}} \ge P_{\mathrm{target}}$, the baseline kernel already saturates the GPU and is launched directly with $W = 1$.
Otherwise, the split count is computed as $W = \min\bigl(\lfloor P_{\mathrm{target}} / P_{\mathrm{base}} \rfloor, T_K\bigr)$ for launching at least $P_{\mathrm{target}}$ thread blocks.

\begin{algorithm}[t]
  \caption{\small \SYSNAME{} Backward --- Setting 1, Compute $dQ_2$ \label{alg:flashkl-bwd1}}
  \begin{algorithmic}[1]
    \small
    \REQUIRE $Q_1, K_1, Q_2, K_2$ in HBM; saved $\mathrm{LSE}_1, \mathrm{LSE}_2\in \mathbb{R}^{N_Q}$ from forward pass; the upstream gradient $dL\in \mathbb{R}^{N_Q}$.
    \FOR{$1 \le i \le T_Q$ \textbf{in parallel}}
      \STATE Load $Q_{1,i}, Q_{2,i}$ and per-row $\mathrm{LSE}_{1}, \mathrm{LSE}_{2}, dL$ from HBM to SRAM.
      \STATE Initialize $dQ_{2,i}^{\mathrm{acc}} \leftarrow (0)_{B_Q \times d_2}$.
      \FOR{$1 \le j \le T_K$}
        \STATE Load $K_{1,j}, K_{2,j}$ from HBM to SRAM.
        \STATE Recompute logits: $S_1 = Q_{1,i}\,K_{1,j}^T$, $S_2 = Q_{2,i}\,K_{2,j}^T$.
        \STATE Recompute probabilities:\\
        $P_1 = e^{S_1 - \mathrm{LSE}_1}$, \quad $P_2 = e^{S_2 - \mathrm{LSE}_2}$.
        \STATE Compute logit gradient: $dS_2 = dL \cdot (P_2 - P_1)$.
        \STATE Accumulate: $dQ_{2,i}^{\mathrm{acc}} \mathrel{+}= dS_2 \, K_{2,j}$.
      \ENDFOR
      \STATE Write $dQ_{2,i}^{\mathrm{acc}}$ to HBM.
    \ENDFOR
  \end{algorithmic}
\end{algorithm}

\section{Backward Pass} \label{sec:bwd}

The backward pass of \SYSNAME{} computes the gradients of the KL divergence with respect to the query and key matrices.
\SYSNAME{} supports two settings corresponding to the two gradient formulas derived in Section~\ref{sec:bg-ad} (Eq.~\ref{eq:kl-case1} and~\ref{eq:kl-case2}).
In both settings, the key design principle is \emph{recomputation}. 
The backward kernels regenerate $P_1$ and $P_2$ tile-by-tile from the saved $\mathrm{LSE}_1$ and $\mathrm{LSE}_2$ vectors, keeping only a $B_Q \times B_K$ logit tile in SRAM at any time.
Unlike the forward pass, the backward kernels do not require online softmax updates, since the global normalization constants $\mathrm{LSE}_1$ and $\mathrm{LSE}_2$ are already known from the forward pass, and attention probabilities can be recovered directly as $P_1^i = e^{S_1^i - \mathrm{LSE}_1}, P_2^i = e^{S_2^i - \mathrm{LSE}_2}$.

\subsection{Algorithm Design}
\subsubsection{Setting 1: Fixed $P_1$, Optimize $P_2$}

In this setting, $P_1$ is the fixed teacher distribution and $P_2$ is the trainable student.
From Eq.~\ref{eq:kl-case1}, the logit-level gradient is $dS_2 = dL \cdot (P_2 - P_1)$, where $dL = \partial \mathcal{L}_{\text{total}} / \partial L$ is the upstream gradient.
The query and key gradients follow as $dQ_2 = dS_2 \, K_2$ and $dK_2 = dS_2^T \, Q_2$.

Algorithm~\ref{alg:flashkl-bwd1} presents the $dQ_2$ computation.
Each thread block loads one query tile and streams all key tiles, recomputing $P_1$ and $P_2$ on the fly and accumulating $dQ_2$ in SRAM registers.
The $dK_2$ computation follows a transposed tiling pattern, where each thread block owns one key tile $K_{2,j}$ and iterates over all query tiles, accumulating $dK_{2,j} = \sum_i dS_{2}^T Q_{2,i}$ in registers before writing the result to HBM.
Since the tiling logic mirrors Algorithm~\ref{alg:flashkl-bwd1} with the roles of $Q$ and $K$ swapped, we omit the pseudocode for brevity.

\begin{algorithm}[t]
  \caption{\small \SYSNAME{} Backward --- Setting 2, Compute $dQ_1$ \label{alg:flashkl-bwd2}}
  \begin{algorithmic}[1]
    \small
    \REQUIRE $Q_1, K_1, Q_2, K_2$ in HBM; saved $\mathrm{LSE}_1, \mathrm{LSE}_2, L\in \mathbb{R}^{N_Q}$ from forward pass; the upstream gradient $dL\in \mathbb{R}^{N_Q}$.
    \FOR{$1 \le i \le T_Q$ \textbf{in parallel}}
      \STATE Load $Q_{1,i}, Q_{2,i}$ and per-row $\mathrm{LSE}_1, \mathrm{LSE}_2, L, dL$ from HBM to SRAM.
      \STATE Precompute $\delta = \mathrm{LSE}_1 - \mathrm{LSE}_2 \in \mathbb{R}^{B_Q}$.
      \STATE Initialize $dQ_{1,i}^{\mathrm{acc}} \leftarrow (0)_{B_Q \times d_1}$.
      \FOR{$1 \le j \le T_K$}
        \STATE Load $K_{1,j}, K_{2,j}$ from HBM to SRAM.
        \STATE Recompute logits: $S_1 = Q_{1,i}\,K_{1,j}^T$, $S_2 = Q_{2,i}\,K_{2,j}^T$.
        \STATE Recompute: $P_1 = e^{S_1 - \mathrm{LSE}_1}$.
        \STATE Compute stable log-ratio: $r = (S_1 - S_2) - \delta$.
        \STATE Compute logit gradient: $dS_1 = dL \cdot P_1 \odot (r - L)$.
        \STATE Accumulate: $dQ_{1,i}^{\mathrm{acc}} \mathrel{+}= dS_1 \, K_{1,j}$.
      \ENDFOR
      \STATE Write $dQ_{1,i}^{\mathrm{acc}}$ to HBM.
    \ENDFOR
  \end{algorithmic}
\end{algorithm}

\subsubsection{Setting 2: Fixed $P_2$, Optimize $P_1$}

In this setting, $P_2$ is the fixed reference and $P_1$ is being optimized.
From Eq.~\ref{eq:kl-case2}, the logit-level gradient is $dS_1 = dL \cdot P_1 \odot (r - L)$, where $r = \log P_1 - \log P_2$ is the element-wise log-ratio and $L = \mathrm{KL}(P_1 \| P_2)$ is the per-row KL value saved from the forward pass.
The query and key gradients also follow as $dQ_1 = dS_1 \, K_1$ and $dK_1 = dS_1^T \, Q_1$.

\paragraph{Numerically stable log-ratio.}
Computing $r^i = \log P_1^i - \log P_2^i$ directly from probabilities is numerically hazardous: when $P^i \approx 0$, taking the logarithm produces $-\infty$ or NaN.
Instead, we express the log-ratio using the saved LSE values:
\begin{equation}
\label{eq:stable-r}
r^i = (S_1^i - S_2^i) - (\mathrm{LSE}_1 - \mathrm{LSE}_2),
\end{equation}
which involves only logit differences and is numerically stable regardless of how small the probabilities are.

Algorithm~\ref{alg:flashkl-bwd2} presents the $dQ_1$ computation.
The structure mirrors Setting 1, with two differences: (i) the kernel additionally loads the forward KL values to compute the baseline-subtracted gradient $r - L$, and (ii) the logit gradient $dS_1 = dL \cdot P_1 \odot (r - L)$ involves an element-wise product with $P_1$ rather than a simple difference.
For $dK_1$, the tiling logic mirrors Algorithm~\ref{alg:flashkl-bwd2} with the roles of $Q$ and $K$ swapped; we omit the pseudocode for brevity.

\begin{figure}[t]
  \centering
  \includegraphics[width=\columnwidth]{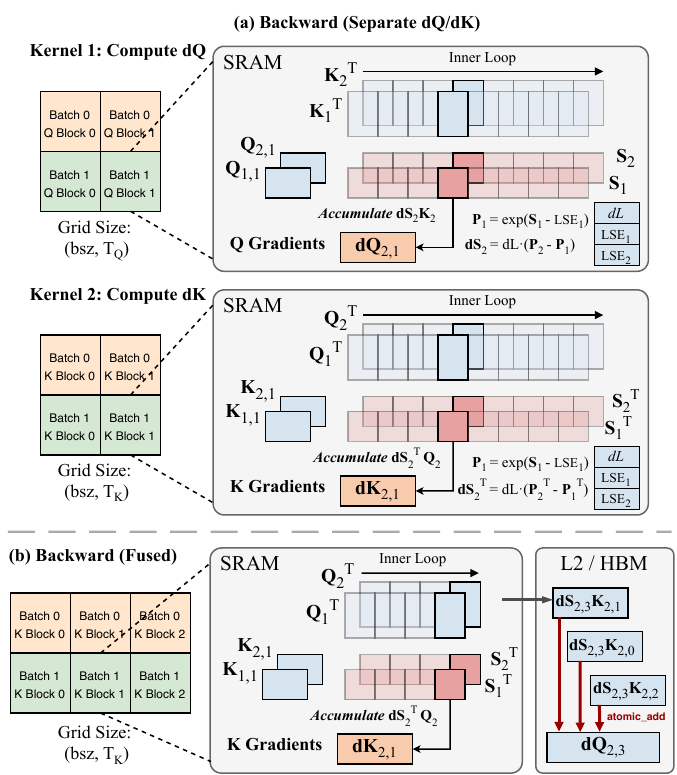}
  \caption{Backward kernel design. (a) Separate strategy: two kernels for $dQ$ and $dK$ on $(\mathrm{bsz}, T_Q)$ and $(\mathrm{bsz}, T_K)$ grids respectively; each owns its output tile but reads inputs and recomputes logit tiles twice. (b) Fused strategy: a single $(\mathrm{bsz}, T_K)$ grid where each thread block owns one $K$ tile, accumulates $dK$ in registers, and writes $dQ$ via \texttt{atomic\_add}.}
  \label{fig:bwd-overview}
\end{figure}

\subsection{GPU Kernel Design} \label{sec:bwd-kernel}

\paragraph{Separate $dQ$/$dK$ Kernels}
As illustrated in Figure~\ref{fig:bwd-overview}(a), a straightforward strategy is to launch two separate kernels for $dQ$ and $dK$, directly instantiating Algorithm~\ref{alg:flashkl-bwd1} for $dQ_2, dK_2$ and Algorithm~\ref{alg:flashkl-bwd2} for $dQ_1, dK_1$.
The $dQ$ kernel is launched with grid $(\mathrm{bsz}, T_Q)$: each thread block owns one query tile, streams all key tiles, and accumulates $dQ$ in registers.
The $dK$ kernel is launched with grid $(\mathrm{bsz}, T_K)$: each thread block owns one key tile, streams all query tiles, and accumulates $dK$ in registers.
Since each kernel owns its output tile exclusively, no cross-block synchronization is needed.

However, this two-kernel strategy incurs redundant HBM traffic. 
Both $Q_1, K_1, Q_2, K_2$ are read from HBM twice (once per kernel), and the logit tiles $S_1, S_2$ are recomputed twice for each tile pair.
A fused single-kernel design would halve these costs by visiting each tile pair exactly once.

\paragraph{Fused Backward Kernel}
Figure~\ref{fig:bwd-overview}(b) illustrates the fused design.
The key obstacle to kernel fusion is a \emph{tiling conflict}: $dQ$ is accumulated along the $K$ dimension, while $dK$ is accumulated along the $Q$ dimension.
Whichever dimension the grid is partitioned on, the other gradient has no exclusively owning thread block and must be written collaboratively via \texttt{atomic\_add}.
For the fused kernel, we partition on the $K$ dimension, launching with grid $(\mathrm{bsz}, T_K)$, since $N_Q \le N_K$ in typical SDPA and the smaller $dQ$ buffer is more likely to remain in L2 cache, thus the cost of concurrent atomic operations is reduced.
Each thread block loads one key tile into registers, iterates over all query tiles in the inner loop and accumulates $dK$ in registers, while per-tile $dQ$ contributions are written via \texttt{atomic\_add} from multiple thread blocks.

\paragraph{Adaptive Strategy Selection}
The fused kernel eliminates redundant HBM reads but introduces atomic contention on $dQ$.
When $N_Q \ll N_K$, the $dQ$ buffer is compact and fits in the L2 cache, making atomic updates inexpensive.
However, when $N_Q$ is large, the $dQ$ working set spills from L2, and contention can negate the HBM savings of fusion.
\SYSNAME{} therefore adaptively selects between the two strategies at runtime: the fused kernel is chosen when $T_Q \times C \le T_K$ (i.e., $dQ$ is small relative to $dK$), where $C$ is an empirically tuned constant; otherwise the separate kernels are used.

\section{Implementation} \label{sec:impl}

We implement all \SYSNAME{} kernels in Triton~\cite{triton} and expose them as PyTorch function modules for drop-in use.
We apply three techniques to maximize the kernel performance.
First, on Hopper and newer architectures, \SYSNAME{} provides a dedicated kernel variant that leverages hardware TMA (Tensor Memory Accelerator) bulk-copy instructions for asynchronous global-to-shared-memory transfers with hardware-accelerated address generation.
When TMA is unavailable, \SYSNAME{} transparently falls back to the generic kernels with standard pointer-based loads.
Second, all exponential operations use \texttt{exp2} ($2^{x \log_2 e}$) instead of the natural exponential, since \texttt{exp2} maps to a single-cycle SFU instruction on NVIDIA GPUs, avoiding the extra FMUL that a natural \texttt{exp} would require.
Third, \SYSNAME{} uses Triton's auto-tuner to select $(B_Q, B_K, \text{num\_warps}, \text{num\_stages})$ from a predefined search space; the auto-tuner profiles each candidate for a given problem shape $(N_Q, N_K, d_1, d_2)$ and caches the fastest configuration for later use.

\section{Evaluation} \label{sec:eval}

\subsection{Setup}

\paragraph{Hardware.}
We evaluate \SYSNAME{} on NVIDIA GPUs spanning two architecture generations: A100 (80\,GB, Ampere) and H200 (141\,GB, Hopper).
The A100 exercises the generic pointer-based kernel path, while the H200 exercises the TMA-accelerated variant.
All experiments on both GPU models use CUDA 12.8, PyTorch 2.8, and Triton 3.4.

\paragraph{Baselines.}
We compare \SYSNAME{} against three implementations of attention distillation:
(i) \textbf{PyTorch}~\cite{pytorch}: the eager implementation that materializes both $N_Q \times N_K$ attention distributions and computes the KL divergence with standard PyTorch operators.
(ii) \textbf{\texttt{torch.compile}}~\cite{torch2}: the PyTorch implementation with graph compilation, which fuses element-wise operations and eliminates intermediate allocations via Inductor-generated Triton kernels, but still materializes the attention distributions.
(iii) \textbf{FLA}~\cite{yang2024fla}: provides a hand-written fused Triton kernel for the KL divergence computation, but the attention distributions are still materialized, with \texttt{torch.compile} enabled.
Since FLA's backward kernel only supports differentiating through the student distribution (Setting~1), we exclude it from Setting~2 comparisons.

\paragraph{Workloads.}
We evaluate diverse workloads by sweeping batch sizes in $\{16, 32\}$ and context lengths from 4K to 512K, under both non-causal and causal masking, and both backward settings.
The head dimension is fixed at $d_1 = d_2 = 128$, a representative value for modern transformer architectures.

\paragraph{Metrics.}
We report the end-to-end latency and peak HBM usage for each implementation.
Since all baselines materialize the $O(N_Q N_K)$ attention distributions, they encounter out of memory (OOM) errors at long contexts.
For these OOM cases, we extrapolate baseline costs from the completed runs using the linear scaling of latency and memory with respect to $\mathrm{bsz} \times N_Q \times N_K$, and denote the extrapolated values with hatched patterns in the figures.

\begin{figure}[t]
  \centering
  \includegraphics[width=\columnwidth]{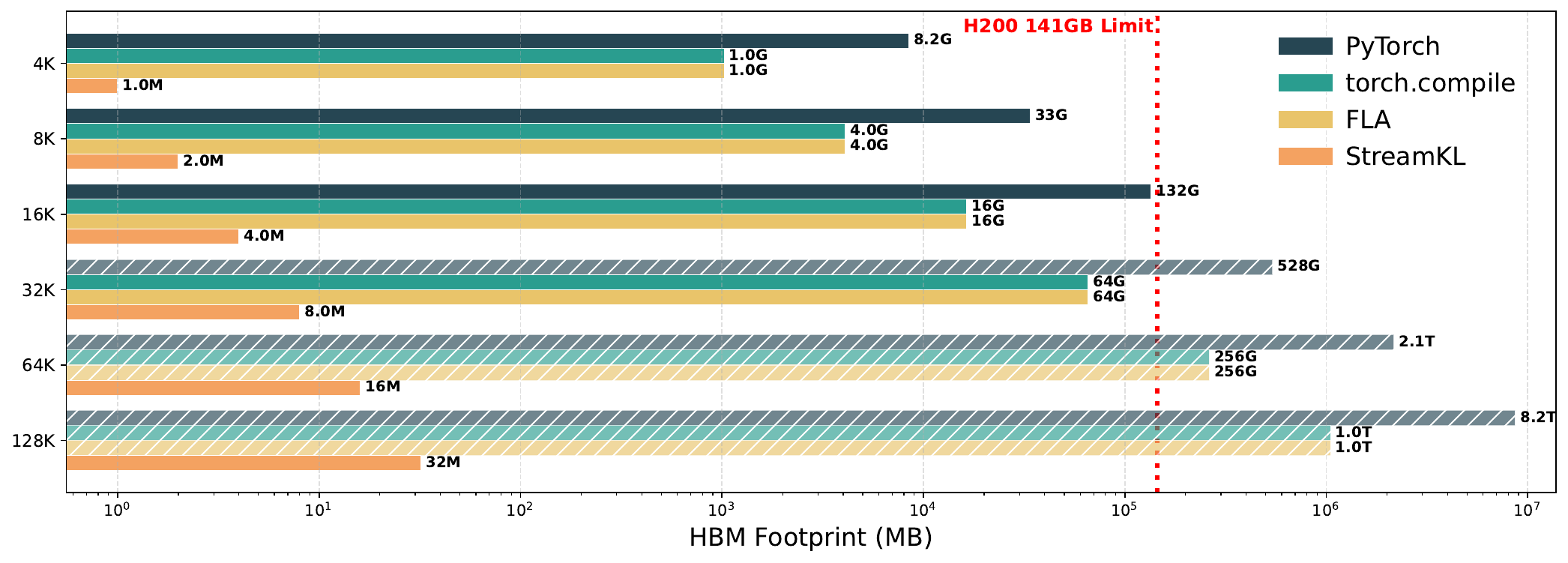}
  \caption{Peak HBM footprint of the non-causal forward pass across context lengths ($N_Q=N_K$, batch 16, log-scale $x$-axis). Hatched bars denote OOM cases with extrapolation.}
  \label{fig:fwd-memory}
\end{figure}

\begin{figure*}[t]
  \centering
  \includegraphics[width=\linewidth]{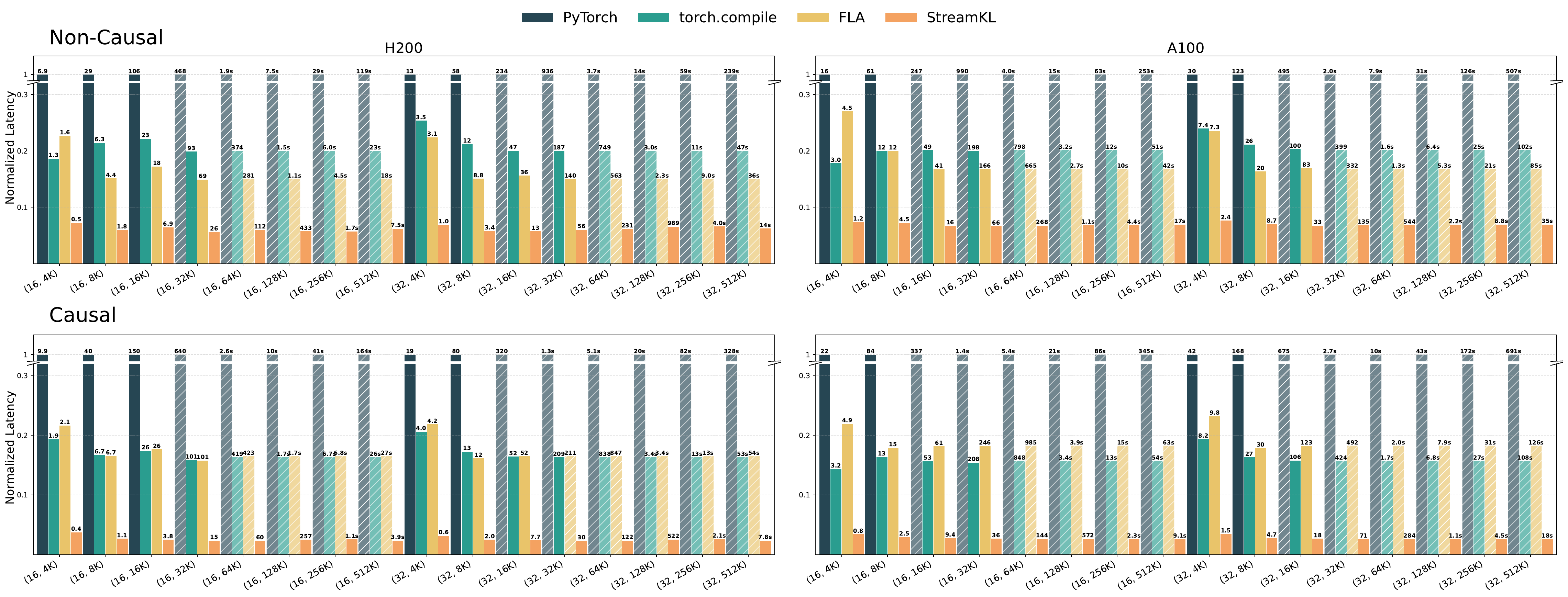}
  \caption{Forward latency on H200 and A100. Each $x$-axis label is a (batch size, $N_Q$) configuration with $N_Q=N_K$. Hatched bars denote OOM cases with extrapolation; numbers are in ms unless suffixed.}
  \label{fig:fwd-latency}
\end{figure*}

\subsection{Forward Pass} \label{sec:eval-fwd}

We compare \SYSNAME{} against the baselines on forward pass peak HBM footprint and latency, sweeping context lengths from 4K to 512K ($N_Q = N_K$) and batch sizes in $\{16, 32\}$ under both causal and non-causal masking.

\paragraph{Memory.}
Figure~\ref{fig:fwd-memory} reports the peak HBM footprint at batch size 16 on H200 across context lengths.
The footprints on A100 follow a similar trend and are omitted for brevity.
As shown, \SYSNAME{}'s footprint stays essentially flat as $N_K$ grows, since it never materializes the $N_Q \times N_K$ attention distributions and only keeps $O(N_Q)$ per-row statistics.
In contrast, all three baselines scale quadratically with the context length.
PyTorch already exceeds the 141\,GB capacity of H200 at 16K, while \texttt{torch.compile} and FLA hit the wall at 32K.
By 64K, \SYSNAME{} consumes over $16{,}000\times$ less HBM than \texttt{torch.compile} and FLA, and over $135{,}000\times$ less than PyTorch, with the gap doubling for every doubling of $N_K$.
FLA consumes about $1.5\times$ more HBM under causal masking than non-causal, since it materializes the causal mask, while \texttt{torch.compile} fuses the masking into the downstream kernel and PyTorch absorbs it into its already-bloated working set.
Only \SYSNAME{} sustains 64K and longer contexts, a regime that is fundamentally out of reach for any baseline.

\paragraph{Latency.}
Figure~\ref{fig:fwd-latency} reports the forward pass latency, normalized to PyTorch with absolute values labeled above each bar.
\SYSNAME{} consistently outperforms all baselines across both GPUs and all inputs, achieving $13.0$--$17.6\times$ speedup over PyTorch, $2.4$--$3.7\times$ over \texttt{torch.compile}, and $2.3$--$3.7\times$ over FLA in the non-causal setting.
The advantage is even more pronounced under causal masking, where \SYSNAME{} achieves $26.6$--$42.7\times$ speedup over PyTorch, $4.2$--$7.0\times$ over \texttt{torch.compile}, and $5.8$--$7.1\times$ over FLA.
This widening gap stems from \SYSNAME{}'s ability to skip masked computation at the kernel level, whereas the baselines still materialize the full $N_Q \times N_K$ attention matrix and apply the mask post hoc, paying even more compute and IO cost than the non-causal case.
The speedups remain stable across context lengths from 4K to 512K, demonstrating that \SYSNAME{} maintains its advantage as the workload scales.

\subsection{Backward Pass} \label{sec:eval-bwd}

\begin{figure}[t]
  \centering
  \includegraphics[width=\columnwidth]{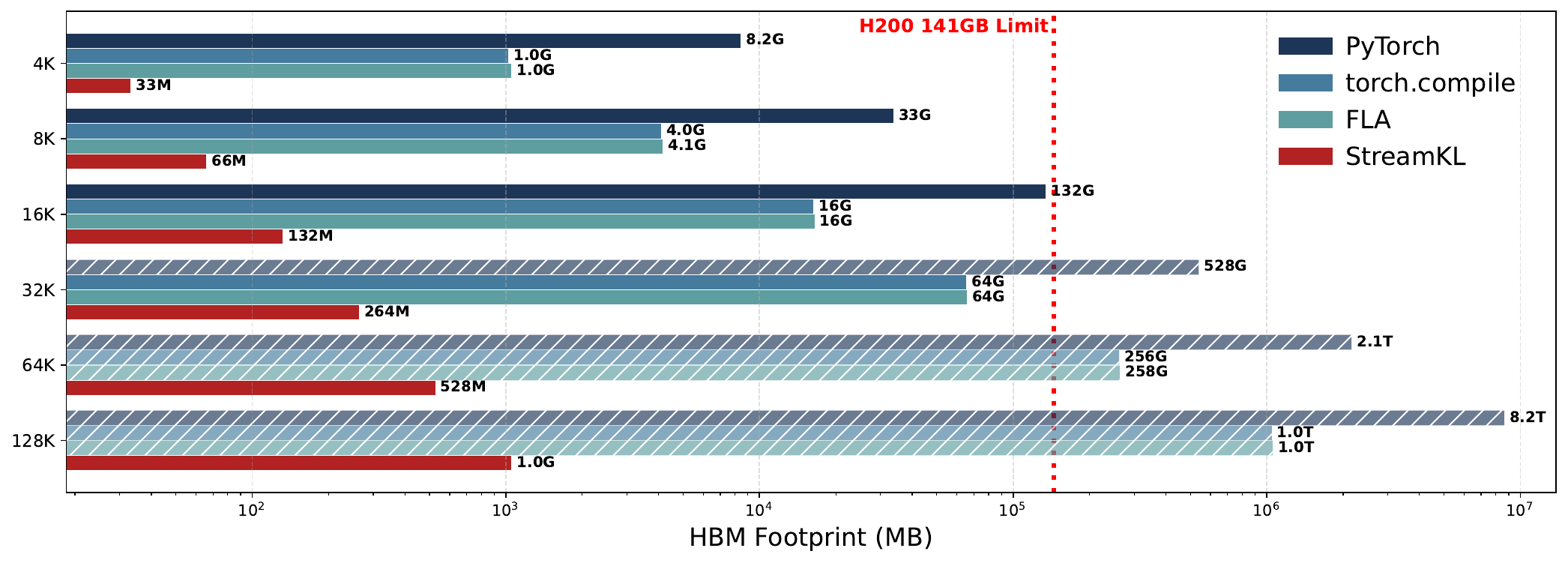}
  \caption{Peak HBM footprint of the non-causal backward pass under Setting 1 across context lengths ($N_Q=N_K$, batch 16). Hatched bars denote OOM cases with extrapolation.}
  \label{fig:bwd-case1-memory}
  \vspace{-0.2cm}
\end{figure}

\begin{figure*}[t]
  \centering
  \includegraphics[width=\linewidth]{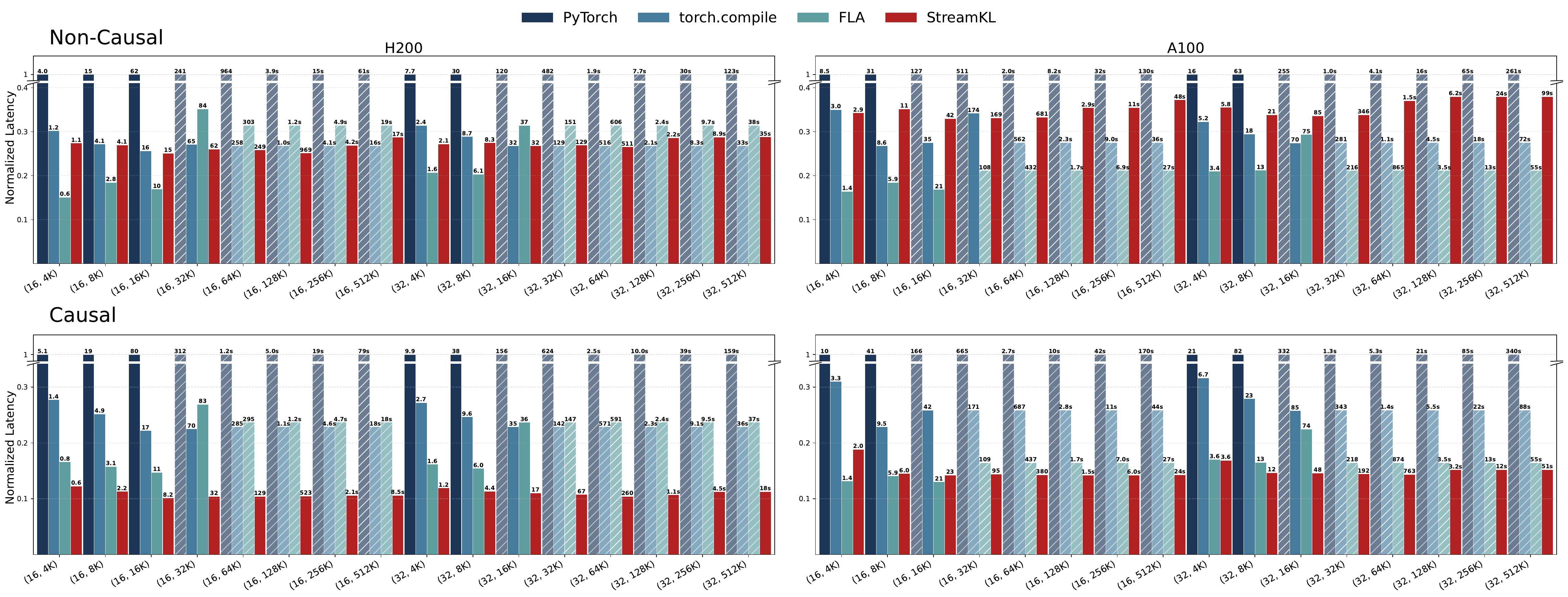}
  \caption{Backward (Setting 1) latency on H200 and A100. Each $x$-axis label is a (batch size, $N_Q$) configuration with $N_Q=N_K$. Hatched bars denote OOM cases with extrapolation; numbers are in ms unless suffixed.}
  \label{fig:bwd-case1-latency}
\end{figure*}

\subsubsection{Setting 1 (Fixed $P_1$, Optimize $P_2$)}

\paragraph{Memory.}
Figure~\ref{fig:bwd-case1-memory} reports the peak HBM footprint of the non-causal backward at batch size 16; the causal numbers are identical and are omitted for brevity.
As shown, \SYSNAME{}'s footprint scales linearly with $N_Q$, growing from 33\,MB at 4K to 4\,GB at 512K, since it stores only the $O((N_Q + N_K) d)$ gradient buffers $dQ_2$, $dK_2$ alongside the per-row LSE statistics needed for recomputation.
The three baselines, in contrast, scale quadratically due to storing $P_1$ and $P_2$, and quickly exhaust the 141\,GB H200.
PyTorch reaches $132$\,GB at 16K and OOMs at 32K, while \texttt{torch.compile} and FLA hit the wall at 64K.
At 64K, \SYSNAME{} consumes over $4{,}000\times$ less HBM than PyTorch and roughly $500\times$ less than \texttt{torch.compile}/FLA, with the gap doubling for every doubling of $N_Q$.

\paragraph{Latency.}
Figure~\ref{fig:bwd-case1-latency} reports the Setting~1 backward latency.
\SYSNAME{} delivers $2.6$--$4.0\times$ speedup over PyTorch in the non-causal setting and $5.3$--$9.9\times$ under causal masking, again driven by the IO and masked computation savings of the fused kernel.
The comparison with \texttt{torch.compile} and FLA, however, is more nuanced.
All baselines store $P_1$ and $P_2$ in HBM after the forward pass and just read them back during the backward pass, while \SYSNAME{} requires recomputation for the $O(1)$ memory footprint.
The cost of this recompute depends on two factors: the GPU's compute throughput and whether causal masking is applied.
On the compute-rich H200, the extra compute is largely hidden by the higher tensor-core throughput, thus the performance of \SYSNAME{} essentially matches \texttt{torch.compile} ($0.93$--$1.16\times$) and stays close to FLA in the non-causal setting; on A100, where compute is the bottleneck, the recompute cost is exposed and \SYSNAME{} trails \texttt{torch.compile} by up to $1.4\times$ and FLA by up to $2.1\times$.
Causal masking, on the other hand, halves the recompute workload since \SYSNAME{} skips masked $QK$ tile computation entirely at the kernel level, flipping the comparison in \SYSNAME{}'s favor on both H200 and A100.
Specifically, on H200 \SYSNAME{} achieves $2.0$--$2.3\times$ speedup over \texttt{torch.compile} and $1.4$--$2.6\times$ over FLA, and on A100 $1.6$--$1.9\times$ and $0.7$--$1.5\times$, respectively.

\begin{figure}[t]
  \centering
  \includegraphics[width=\columnwidth]{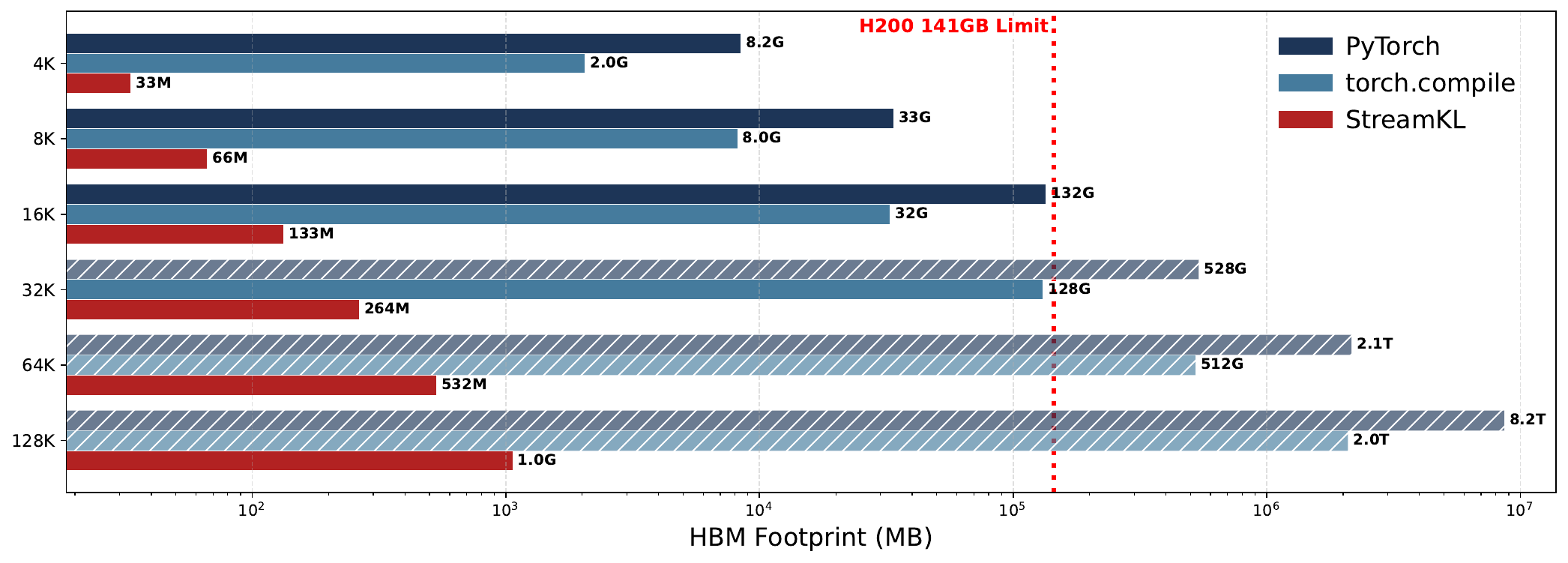}
  \caption{Peak HBM footprint of the non-causal backward pass under Setting 2 across context lengths ($N_Q=N_K$, batch 16). Hatched bars denote OOM cases with extrapolation.}
  \label{fig:bwd-case2-memory}
\end{figure}

\begin{figure*}[t]
  \centering
  \includegraphics[width=\linewidth]{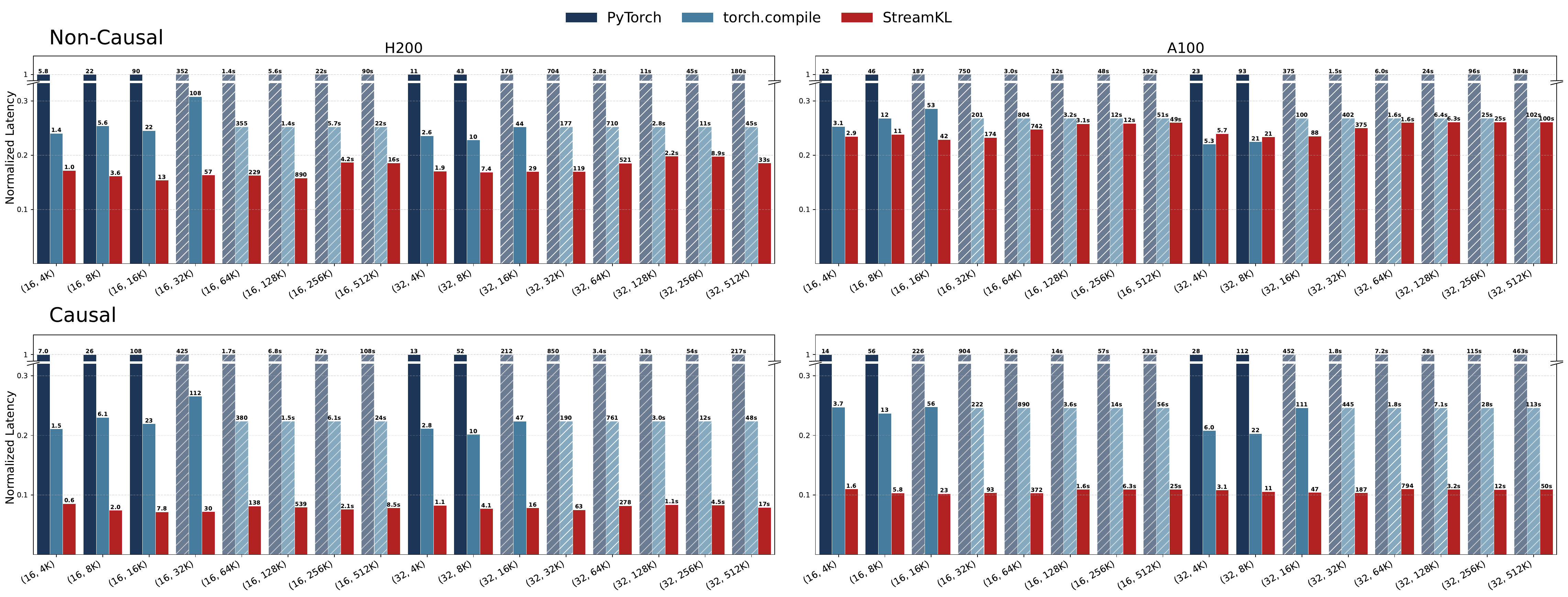}
  \caption{Backward (Setting 2) latency on H200 and A100. Each $x$-axis label is a (batch size, $N_Q$) configuration with $N_Q=N_K$. Hatched bars denote OOM cases with extrapolation; numbers are in ms unless suffixed.}
  \label{fig:bwd-case2-latency}
\end{figure*}

\subsubsection{Setting 2 (Fixed $P_2$, Optimize $P_1$)}

\paragraph{Memory.}
Figure~\ref{fig:bwd-case2-memory} reports the peak HBM footprint of the non-causal Setting~2 backward at batch size 16 on H200; as in Setting~1, the causal numbers are identical and are omitted for brevity.
\SYSNAME{} retains the same linear $O(N_Q)$ scaling and the same absolute footprint as in Setting~1, since the only addition is the per-row KL scalar $L$ used by the kernel.
However, \texttt{torch.compile} consumes roughly $2\times$ more HBM than in Setting~1 because the autograd tape now has to save additional $O(N_Q N_K)$ intermediates ($P_1$ and $\log P_1$), and the log-ratio $r$ to differentiate through every occurrence of $P_1$ in $P_1 \odot (r - L)$.
PyTorch's footprint is essentially unchanged from Setting~1, since eager mode already materializes all intermediates regardless of which branch carries gradients.

\paragraph{Latency.}
Figure~\ref{fig:bwd-case2-latency} reports the Setting~2 backward latency. 
\SYSNAME{} delivers $3.8$--$6.5\times$ speedup over PyTorch in the non-causal setting and $9.1$--$14.0\times$ under causal masking, mirroring the forward-pass trend.
The comparison with \texttt{torch.compile} now favors \SYSNAME{} more decisively than in Setting~1: even in the recompute-exposed non-causal regime \SYSNAME{} achieves $1.27$--$1.89\times$ speedup on H200 and $0.92$--$1.25\times$ on A100, whereas Setting~1 had \SYSNAME{} only tied with \texttt{torch.compile}.
The reason is that the Setting~2 gradient $dS_1 = P_1 \odot (r - L)$ forces the autograd tape to differentiate through $P_1$ at three locations (the $\exp$, the multiplier, and the log-difference), producing a longer and more expensive compiled graph than Setting~1's simple $dS_2 = P_2 - P_1$.
In contrast, the fused backward kernel of \SYSNAME{} only adds a per-row scalar load and a handful of ALU instructions over its Setting~1 counterpart.
Causal masking again halves the recompute workload and lifts \SYSNAME{}'s lead over \texttt{torch.compile} to $2.5$--$3.7\times$ on H200 and $1.9$--$2.4\times$ on A100.

\subsection{Ablation Study}

\begin{figure}[t]
  \centering
  \includegraphics[width=\columnwidth]{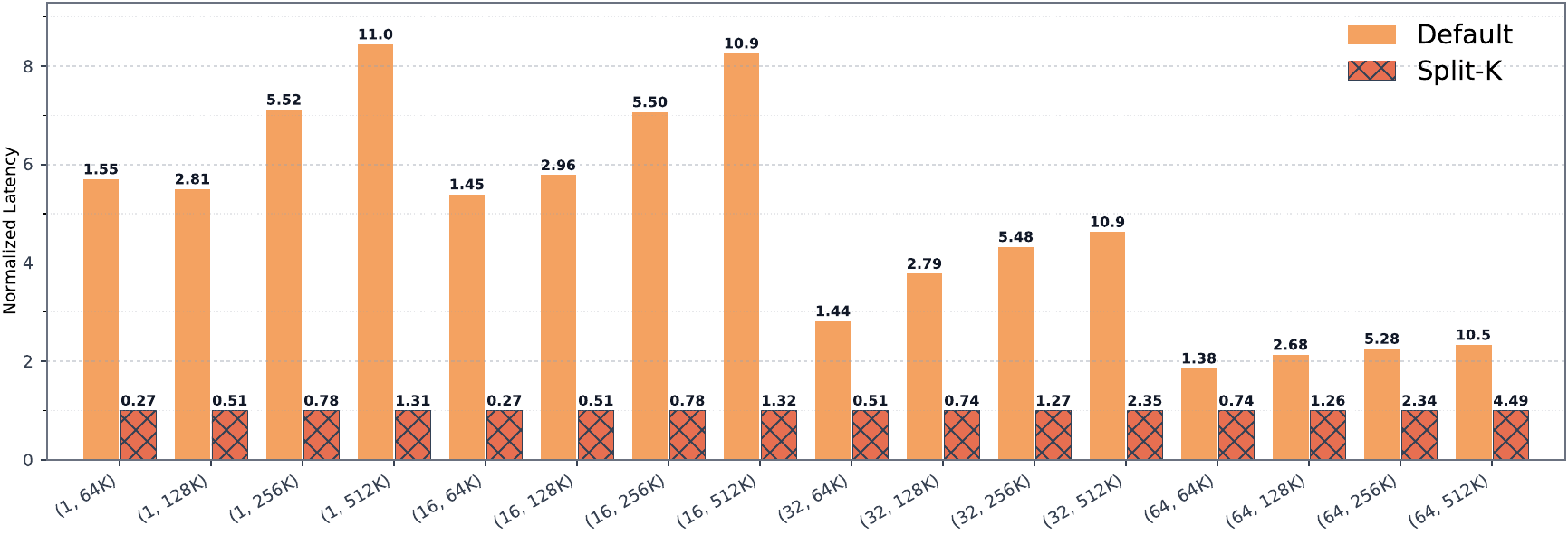}
  \caption{Split-K vs.\ default forward kernel (non-causal, batch size 16). Each $x$-axis label is a ($N_Q$,$N_K$) configuration. Numbers are absolute latencies in ms.}
  \label{fig:fwd-ablation-non-causal}
\end{figure}

\subsubsection{Split-K Variant for Forward}

Figure~\ref{fig:fwd-ablation-non-causal} compares the default forward kernel against the split-K variant from Sec.~\ref{sec:fwd-kernel} on inputs with small $N_Q$, which cannot saturate all SMs in the default kernel.
The split-K variant delivers $5.4$--$8.4\times$ speedup at $N_Q = 1$ and $16$, $2.8$--$4.6\times$ at $N_Q = 32$, and $1.9$--$2.3\times$ at $N_Q = 64$, with the gain widening as $N_K$ grows.
The speedup shrinks as $N_Q$ grows since the default kernel launches enough query blocks to keep SMs fully utilized, leaving no idle parallelism for the split-K variant.
The causal setting exhibits the same trend and is omitted for brevity.

\begin{figure}[t]
  \centering
  \includegraphics[width=\columnwidth]{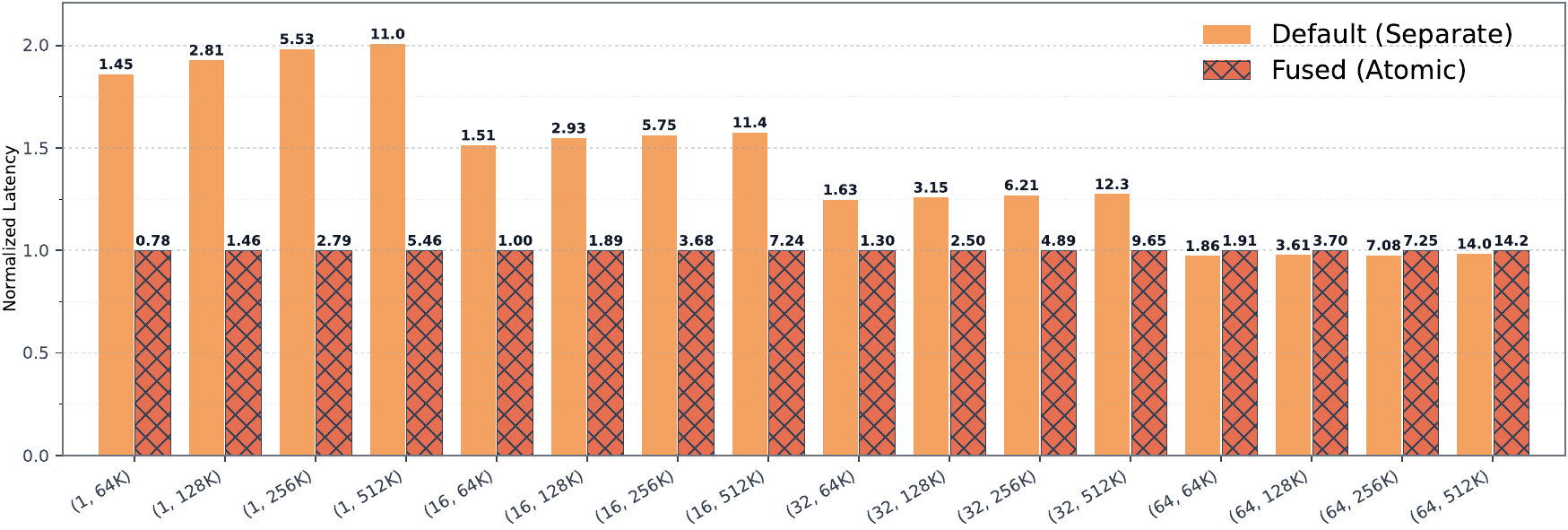}
  \caption{Separate vs.\ fused backward kernel (non-causal, Setting~1, batch size 16). Each $x$-axis label is a ($N_Q$,$N_K$) configuration. Numbers are absolute latencies in ms.}
  \label{fig:bwd-case1-ablation-non-causal}
  \vspace{-0.1cm}
\end{figure}

\subsubsection{Separate/Fused Kernels for Backward}

Figure~\ref{fig:bwd-case1-ablation-non-causal} compares the two backward strategies from Sec.~\ref{sec:bwd-kernel}: the separate $dQ$/$dK$ kernels that read inputs and recompute logits twice, versus the fused atomic kernel that halves both the HBM traffic and the logit recomputation at the cost of \texttt{atomic\_add} contention on $dQ$.
The fused atomic kernel is $2.0\times$ faster than the separate kernels at $N_Q = 1$, $1.5\times$--$1.6\times$ faster at $N_Q = 16$, and $1.2\times$--$1.3\times$ faster at $N_Q = 32$, as the small $dQ$ buffer stays L2-resident and absorbs atomic updates without round trips to HBM.
At $N_Q = 64$, the $dQ$ buffer spills from L2 and atomic contention overtakes the savings, flipping the comparison so that the separate kernels are marginally ($1.02\times$) faster.
This crossover justifies the adaptive runtime selection in Sec.~\ref{sec:bwd-kernel}, which picks the fused kernel only when $dQ$ is small enough to stay L2-resident, and falls back to the separate kernels otherwise.
Setting~2 and the causal cases exhibit the same trend and are omitted.

\begin{table}[t]
  \centering
  \footnotesize
  \setlength{\abovecaptionskip}{3pt}
  \caption{TMA-enabled vs non-TMA speedup on H200.}
  \label{tab:sm90-speedup}
  \begin{tabular}{l c c c c c c}
    \toprule
    & \multicolumn{3}{c}{Non-causal} & \multicolumn{3}{c}{Causal} \\
    \cmidrule(lr){2-4} \cmidrule(lr){5-7}
    Pass & min & median & max & min & median & max \\
    \midrule
    Forward            & 0.97 & 1.02 & 1.11 & 0.95 & 1.02 & 1.04 \\
    Backward Setting~1 & 0.99 & 1.03 & 1.07 & 1.05 & 1.08 & 1.12 \\
    Backward Setting~2 & 0.95 & 1.07 & 1.12 & 1.01 & 1.10 & 1.18 \\
    \bottomrule
  \end{tabular}
\end{table}

\subsubsection{Impact of TMA}

Table~\ref{tab:sm90-speedup} reports the speedup of the TMA-enabled kernels over the pointer-based fallback on H200.
Enabling TMA yields up to $1.11\times$ on the forward pass, $1.08$--$1.12\times$ on the Setting~1 backward, and $1.07$--$1.18\times$ on the Setting~2 backward, with the largest gains on the causal variants.
The backward passes benefit more because they issue twice as many global-to-shared loads per tile, allowing asynchronous bulk copies to hide a larger fraction of memory latency.
Setting~2 benefits the most since TMA's asynchronous loads overlap with its extra log-ratio arithmetic.
Overall, the TMA implementation is consistently faster and no more than a few percent slower at worst.

\section{Conclusion}

We presented \SYSNAME{}, the first fused, one-pass primitive for attention distillation that reduces the extra HBM footprint from $O(N_Q N_K)$ to $O(1)$. 
Through a novel online formulation of the coupled two-distribution KL reduction, \SYSNAME{} computes both forward and backward passes in tiled kernels that maintain only per-row statistics on chip. 
Experiments on A100 and H200 GPUs show consistent speedups and memory savings, enabling efficient attention distillation at context lengths of 64K and beyond on a single GPU.

\clearpage

\bibliographystyle{ACM-Reference-Format}
\bibliography{ref}


\begin{thebibliography}{21}


\ifx \showCODEN    \undefined \def \showCODEN     #1{\unskip}     \fi
\ifx \showISBNx    \undefined \def \showISBNx     #1{\unskip}     \fi
\ifx \showISBNxiii \undefined \def \showISBNxiii  #1{\unskip}     \fi
\ifx \showISSN     \undefined \def \showISSN      #1{\unskip}     \fi
\ifx \showLCCN     \undefined \def \showLCCN      #1{\unskip}     \fi
\ifx \shownote     \undefined \def \shownote      #1{#1}          \fi
\ifx \showarticletitle \undefined \def \showarticletitle #1{#1}   \fi
\ifx \showURL      \undefined \def \showURL       {\relax}        \fi
\providecommand\bibfield[2]{#2}
\providecommand\bibinfo[2]{#2}
\providecommand\natexlab[1]{#1}
\providecommand\showeprint[2][]{arXiv:#2}

\bibitem[Agrawal et~al\mbox{.}(2025)]%
        {agrawal-etal-2025-multilingual}
\bibfield{author}{\bibinfo{person}{Sanjay Agrawal}, \bibinfo{person}{Deep
  Nayak}, {and} \bibinfo{person}{Vivek~Varadarajan Sembium}.}
  \bibinfo{year}{2025}\natexlab{}.
\newblock \showarticletitle{Multilingual Continual Learning using Attention
  Distillation}. In \bibinfo{booktitle}{\emph{Proceedings of the 31st
  International Conference on Computational Linguistics: Industry Track}},
  \bibfield{editor}{\bibinfo{person}{Owen Rambow}, \bibinfo{person}{Leo
  Wanner}, \bibinfo{person}{Marianna Apidianaki}, \bibinfo{person}{Hend
  Al-Khalifa}, \bibinfo{person}{Barbara~Di Eugenio}, \bibinfo{person}{Steven
  Schockaert}, \bibinfo{person}{Kareem Darwish}, {and} \bibinfo{person}{Apoorv
  Agarwal}} (Eds.). \bibinfo{publisher}{Association for Computational
  Linguistics}, \bibinfo{address}{Abu Dhabi, UAE}, \bibinfo{pages}{91--99}.
\newblock
\urldef\tempurl%
\url{https://aclanthology.org/2025.coling-industry.8/}
\showURL{%
\tempurl}


\bibitem[Ansel et~al\mbox{.}(2024)]%
        {torch2}
\bibfield{author}{\bibinfo{person}{Jason Ansel}, \bibinfo{person}{Edward Yang},
  \bibinfo{person}{Horace He}, \bibinfo{person}{Natalia Gimelshein},
  \bibinfo{person}{Animesh Jain}, \bibinfo{person}{Michael Voznesensky},
  \bibinfo{person}{Bin Bao}, \bibinfo{person}{Peter Bell},
  \bibinfo{person}{David Berard}, \bibinfo{person}{Evgeni Burovski},
  \bibinfo{person}{Geeta Chauhan}, \bibinfo{person}{Anjali Chourdia},
  \bibinfo{person}{Will Constable}, \bibinfo{person}{Alban Desmaison},
  \bibinfo{person}{Zachary DeVito}, \bibinfo{person}{Elias Ellison},
  \bibinfo{person}{Will Feng}, \bibinfo{person}{Jiong Gong},
  \bibinfo{person}{Michael Gschwind}, \bibinfo{person}{Brian Hirsh},
  \bibinfo{person}{Sherlock Huang}, \bibinfo{person}{Kshiteej Kalambarkar},
  \bibinfo{person}{Laurent Kirsch}, \bibinfo{person}{Michael Lazos},
  \bibinfo{person}{Mario Lezcano}, \bibinfo{person}{Yanbo Liang},
  \bibinfo{person}{Jason Liang}, \bibinfo{person}{Yinghai Lu},
  \bibinfo{person}{C.~K. Luk}, \bibinfo{person}{Bert Maher},
  \bibinfo{person}{Yunjie Pan}, \bibinfo{person}{Christian Puhrsch},
  \bibinfo{person}{Matthias Reso}, \bibinfo{person}{Mark Saroufim},
  \bibinfo{person}{Marcos~Yukio Siraichi}, \bibinfo{person}{Helen Suk},
  \bibinfo{person}{Shunting Zhang}, \bibinfo{person}{Michael Suo},
  \bibinfo{person}{Phil Tillet}, \bibinfo{person}{Xu Zhao},
  \bibinfo{person}{Eikan Wang}, \bibinfo{person}{Keren Zhou},
  \bibinfo{person}{Richard Zou}, \bibinfo{person}{Xiaodong Wang},
  \bibinfo{person}{Ajit Mathews}, \bibinfo{person}{William Wen},
  \bibinfo{person}{Gregory Chanan}, \bibinfo{person}{Peng Wu}, {and}
  \bibinfo{person}{Soumith Chintala}.} \bibinfo{year}{2024}\natexlab{}.
\newblock \showarticletitle{PyTorch 2: Faster Machine Learning Through Dynamic
  Python Bytecode Transformation and Graph Compilation}. In
  \bibinfo{booktitle}{\emph{Proceedings of the 29th ACM International
  Conference on Architectural Support for Programming Languages and Operating
  Systems, Volume 2}} (La Jolla, CA, USA) \emph{(\bibinfo{series}{ASPLOS
  '24})}. \bibinfo{publisher}{Association for Computing Machinery},
  \bibinfo{address}{New York, NY, USA}, \bibinfo{pages}{929--947}.
\newblock
\showISBNx{9798400703850}
\href{https://doi.org/10.1145/3620665.3640366}{doi:\nolinkurl{10.1145/3620665.3640366}}


\bibitem[Brown et~al\mbox{.}(2020)]%
        {gpt}
\bibfield{author}{\bibinfo{person}{Tom~B. Brown}, \bibinfo{person}{Benjamin
  Mann}, \bibinfo{person}{Nick Ryder}, \bibinfo{person}{Melanie Subbiah},
  \bibinfo{person}{Jared Kaplan}, \bibinfo{person}{Prafulla Dhariwal},
  \bibinfo{person}{Arvind Neelakantan}, \bibinfo{person}{Pranav Shyam},
  \bibinfo{person}{Girish Sastry}, \bibinfo{person}{Amanda Askell},
  \bibinfo{person}{Sandhini Agarwal}, \bibinfo{person}{Ariel Herbert-Voss},
  \bibinfo{person}{Gretchen Krueger}, \bibinfo{person}{Tom Henighan},
  \bibinfo{person}{Rewon Child}, \bibinfo{person}{Aditya Ramesh},
  \bibinfo{person}{Daniel~M. Ziegler}, \bibinfo{person}{Jeffrey Wu},
  \bibinfo{person}{Clemens Winter}, \bibinfo{person}{Christopher Hesse},
  \bibinfo{person}{Mark Chen}, \bibinfo{person}{Eric Sigler},
  \bibinfo{person}{Mateusz Litwin}, \bibinfo{person}{Scott Gray},
  \bibinfo{person}{Benjamin Chess}, \bibinfo{person}{Jack Clark},
  \bibinfo{person}{Christopher Berner}, \bibinfo{person}{Sam McCandlish},
  \bibinfo{person}{Alec Radford}, \bibinfo{person}{Ilya Sutskever}, {and}
  \bibinfo{person}{Dario Amodei}.} \bibinfo{year}{2020}\natexlab{}.
\newblock \bibinfo{title}{Language Models are Few-Shot Learners}.
\newblock
\showeprint[arxiv]{2005.14165}~[cs.CL]
\urldef\tempurl%
\url{https://arxiv.org/abs/2005.14165}
\showURL{%
\tempurl}


\bibitem[Choi et~al\mbox{.}(2025)]%
        {miniq}
\bibfield{author}{\bibinfo{person}{Kanghyun Choi}, \bibinfo{person}{Hyeyoon
  Lee}, \bibinfo{person}{Dain Kwon}, \bibinfo{person}{SunJong Park},
  \bibinfo{person}{Kyuyeun Kim}, \bibinfo{person}{Noseong Park},
  \bibinfo{person}{Jonghyun Choi}, {and} \bibinfo{person}{Jinho Lee}.}
  \bibinfo{year}{2025}\natexlab{}.
\newblock \showarticletitle{MimiQ: Low-Bit Data-Free Quantization of Vision
  Transformers with Encouraging Inter-Head Attention Similarity}.
\newblock \bibinfo{journal}{\emph{Proceedings of the AAAI Conference on
  Artificial Intelligence}} \bibinfo{volume}{39}, \bibinfo{number}{15}
  (\bibinfo{date}{April} \bibinfo{year}{2025}), \bibinfo{pages}{16037–16045}.
\newblock
\showISSN{2159-5399}
\href{https://doi.org/10.1609/aaai.v39i15.33761}{doi:\nolinkurl{10.1609/aaai.v39i15.33761}}


\bibitem[Dao(2023)]%
        {flash-attn2}
\bibfield{author}{\bibinfo{person}{Tri Dao}.} \bibinfo{year}{2023}\natexlab{}.
\newblock \bibinfo{title}{FlashAttention-2: Faster Attention with Better
  Parallelism and Work Partitioning}.
\newblock
\showeprint[arxiv]{2307.08691}~[cs.LG]
\urldef\tempurl%
\url{https://arxiv.org/abs/2307.08691}
\showURL{%
\tempurl}


\bibitem[Dao et~al\mbox{.}(2022)]%
        {flash-attention}
\bibfield{author}{\bibinfo{person}{Tri Dao}, \bibinfo{person}{Daniel~Y. Fu},
  \bibinfo{person}{Stefano Ermon}, \bibinfo{person}{Atri Rudra}, {and}
  \bibinfo{person}{Christopher R{\'e}}.} \bibinfo{year}{2022}\natexlab{}.
\newblock \showarticletitle{Flash{A}ttention: Fast and Memory-Efficient Exact
  Attention with {IO}-Awareness}. In \bibinfo{booktitle}{\emph{Advances in
  Neural Information Processing Systems}}, Vol.~\bibinfo{volume}{35}.
\newblock


\bibitem[DeepSeek-AI et~al\mbox{.}(2025)]%
        {deepseek-v32}
\bibfield{author}{\bibinfo{person}{DeepSeek-AI}, \bibinfo{person}{Aixin Liu},
  \bibinfo{person}{Aoxue Mei}, \bibinfo{person}{Bangcai Lin},
  \bibinfo{person}{Bing Xue}, \bibinfo{person}{Bingxuan Wang},
  \bibinfo{person}{Bingzheng Xu}, \bibinfo{person}{Bochao Wu},
  \bibinfo{person}{Bowei Zhang}, \bibinfo{person}{Chaofan Lin},
  \bibinfo{person}{Chen Dong}, \bibinfo{person}{Chengda Lu},
  \bibinfo{person}{Chenggang Zhao}, \bibinfo{person}{Chengqi Deng},
  \bibinfo{person}{Chenhao Xu}, \bibinfo{person}{Chong Ruan},
  \bibinfo{person}{Damai Dai}, \bibinfo{person}{Daya Guo},
  \bibinfo{person}{Dejian Yang}, \bibinfo{person}{Deli Chen},
  \bibinfo{person}{Erhang Li}, \bibinfo{person}{Fangqi Zhou},
  \bibinfo{person}{Fangyun Lin}, \bibinfo{person}{Fucong Dai},
  \bibinfo{person}{Guangbo Hao}, \bibinfo{person}{Guanting Chen},
  \bibinfo{person}{Guowei Li}, \bibinfo{person}{H. Zhang},
  \bibinfo{person}{Hanwei Xu}, \bibinfo{person}{Hao Li},
  \bibinfo{person}{Haofen Liang}, \bibinfo{person}{Haoran Wei},
  \bibinfo{person}{Haowei Zhang}, \bibinfo{person}{Haowen Luo},
  \bibinfo{person}{Haozhe Ji}, \bibinfo{person}{Honghui Ding},
  \bibinfo{person}{Hongxuan Tang}, \bibinfo{person}{Huanqi Cao},
  \bibinfo{person}{Huazuo Gao}, \bibinfo{person}{Hui Qu}, \bibinfo{person}{Hui
  Zeng}, \bibinfo{person}{Jialiang Huang}, \bibinfo{person}{Jiashi Li},
  \bibinfo{person}{Jiaxin Xu}, \bibinfo{person}{Jiewen Hu},
  \bibinfo{person}{Jingchang Chen}, \bibinfo{person}{Jingting Xiang},
  \bibinfo{person}{Jingyang Yuan}, \bibinfo{person}{Jingyuan Cheng},
  \bibinfo{person}{Jinhua Zhu}, \bibinfo{person}{Jun Ran},
  \bibinfo{person}{Junguang Jiang}, \bibinfo{person}{Junjie Qiu},
  \bibinfo{person}{Junlong Li}, \bibinfo{person}{Junxiao Song},
  \bibinfo{person}{Kai Dong}, \bibinfo{person}{Kaige Gao},
  \bibinfo{person}{Kang Guan}, \bibinfo{person}{Kexin Huang},
  \bibinfo{person}{Kexing Zhou}, \bibinfo{person}{Kezhao Huang},
  \bibinfo{person}{Kuai Yu}, \bibinfo{person}{Lean Wang},
  \bibinfo{person}{Lecong Zhang}, \bibinfo{person}{Lei Wang},
  \bibinfo{person}{Liang Zhao}, \bibinfo{person}{Liangsheng Yin},
  \bibinfo{person}{Lihua Guo}, \bibinfo{person}{Lingxiao Luo},
  \bibinfo{person}{Linwang Ma}, \bibinfo{person}{Litong Wang},
  \bibinfo{person}{Liyue Zhang}, \bibinfo{person}{M.~S. Di},
  \bibinfo{person}{M.~Y Xu}, \bibinfo{person}{Mingchuan Zhang},
  \bibinfo{person}{Minghua Zhang}, \bibinfo{person}{Minghui Tang},
  \bibinfo{person}{Mingxu Zhou}, \bibinfo{person}{Panpan Huang},
  \bibinfo{person}{Peixin Cong}, \bibinfo{person}{Peiyi Wang},
  \bibinfo{person}{Qiancheng Wang}, \bibinfo{person}{Qihao Zhu},
  \bibinfo{person}{Qingyang Li}, \bibinfo{person}{Qinyu Chen},
  \bibinfo{person}{Qiushi Du}, \bibinfo{person}{Ruiling Xu},
  \bibinfo{person}{Ruiqi Ge}, \bibinfo{person}{Ruisong Zhang},
  \bibinfo{person}{Ruizhe Pan}, \bibinfo{person}{Runji Wang},
  \bibinfo{person}{Runqiu Yin}, \bibinfo{person}{Runxin Xu},
  \bibinfo{person}{Ruomeng Shen}, \bibinfo{person}{Ruoyu Zhang},
  \bibinfo{person}{S.~H. Liu}, \bibinfo{person}{Shanghao Lu},
  \bibinfo{person}{Shangyan Zhou}, \bibinfo{person}{Shanhuang Chen},
  \bibinfo{person}{Shaofei Cai}, \bibinfo{person}{Shaoyuan Chen},
  \bibinfo{person}{Shengding Hu}, \bibinfo{person}{Shengyu Liu},
  \bibinfo{person}{Shiqiang Hu}, \bibinfo{person}{Shirong Ma},
  \bibinfo{person}{Shiyu Wang}, \bibinfo{person}{Shuiping Yu},
  \bibinfo{person}{Shunfeng Zhou}, \bibinfo{person}{Shuting Pan},
  \bibinfo{person}{Songyang Zhou}, \bibinfo{person}{Tao Ni},
  \bibinfo{person}{Tao Yun}, \bibinfo{person}{Tian Pei}, \bibinfo{person}{Tian
  Ye}, \bibinfo{person}{Tianyuan Yue}, \bibinfo{person}{Wangding Zeng},
  \bibinfo{person}{Wen Liu}, \bibinfo{person}{Wenfeng Liang},
  \bibinfo{person}{Wenjie Pang}, \bibinfo{person}{Wenjing Luo},
  \bibinfo{person}{Wenjun Gao}, \bibinfo{person}{Wentao Zhang},
  \bibinfo{person}{Xi Gao}, \bibinfo{person}{Xiangwen Wang},
  \bibinfo{person}{Xiao Bi}, \bibinfo{person}{Xiaodong Liu},
  \bibinfo{person}{Xiaohan Wang}, \bibinfo{person}{Xiaokang Chen},
  \bibinfo{person}{Xiaokang Zhang}, \bibinfo{person}{Xiaotao Nie},
  \bibinfo{person}{Xin Cheng}, \bibinfo{person}{Xin Liu}, \bibinfo{person}{Xin
  Xie}, \bibinfo{person}{Xingchao Liu}, \bibinfo{person}{Xingkai Yu},
  \bibinfo{person}{Xingyou Li}, \bibinfo{person}{Xinyu Yang},
  \bibinfo{person}{Xinyuan Li}, \bibinfo{person}{Xu Chen},
  \bibinfo{person}{Xuecheng Su}, \bibinfo{person}{Xuehai Pan},
  \bibinfo{person}{Xuheng Lin}, \bibinfo{person}{Xuwei Fu},
  \bibinfo{person}{Y.~Q. Wang}, \bibinfo{person}{Yang Zhang},
  \bibinfo{person}{Yanhong Xu}, \bibinfo{person}{Yanru Ma},
  \bibinfo{person}{Yao Li}, \bibinfo{person}{Yao Li}, \bibinfo{person}{Yao
  Zhao}, \bibinfo{person}{Yaofeng Sun}, \bibinfo{person}{Yaohui Wang},
  \bibinfo{person}{Yi Qian}, \bibinfo{person}{Yi Yu}, \bibinfo{person}{Yichao
  Zhang}, \bibinfo{person}{Yifan Ding}, \bibinfo{person}{Yifan Shi},
  \bibinfo{person}{Yiliang Xiong}, \bibinfo{person}{Ying He},
  \bibinfo{person}{Ying Zhou}, \bibinfo{person}{Yinmin Zhong},
  \bibinfo{person}{Yishi Piao}, \bibinfo{person}{Yisong Wang},
  \bibinfo{person}{Yixiao Chen}, \bibinfo{person}{Yixuan Tan},
  \bibinfo{person}{Yixuan Wei}, \bibinfo{person}{Yiyang Ma},
  \bibinfo{person}{Yiyuan Liu}, \bibinfo{person}{Yonglun Yang},
  \bibinfo{person}{Yongqiang Guo}, \bibinfo{person}{Yongtong Wu},
  \bibinfo{person}{Yu Wu}, \bibinfo{person}{Yuan Cheng}, \bibinfo{person}{Yuan
  Ou}, \bibinfo{person}{Yuanfan Xu}, \bibinfo{person}{Yuduan Wang},
  \bibinfo{person}{Yue Gong}, \bibinfo{person}{Yuhan Wu},
  \bibinfo{person}{Yuheng Zou}, \bibinfo{person}{Yukun Li},
  \bibinfo{person}{Yunfan Xiong}, \bibinfo{person}{Yuxiang Luo},
  \bibinfo{person}{Yuxiang You}, \bibinfo{person}{Yuxuan Liu},
  \bibinfo{person}{Yuyang Zhou}, \bibinfo{person}{Z.~F. Wu},
  \bibinfo{person}{Z.~Z. Ren}, \bibinfo{person}{Zehua Zhao},
  \bibinfo{person}{Zehui Ren}, \bibinfo{person}{Zhangli Sha},
  \bibinfo{person}{Zhe Fu}, \bibinfo{person}{Zhean Xu}, \bibinfo{person}{Zhenda
  Xie}, \bibinfo{person}{Zhengyan Zhang}, \bibinfo{person}{Zhewen Hao},
  \bibinfo{person}{Zhibin Gou}, \bibinfo{person}{Zhicheng Ma},
  \bibinfo{person}{Zhigang Yan}, \bibinfo{person}{Zhihong Shao},
  \bibinfo{person}{Zhixian Huang}, \bibinfo{person}{Zhiyu Wu},
  \bibinfo{person}{Zhuoshu Li}, \bibinfo{person}{Zhuping Zhang},
  \bibinfo{person}{Zian Xu}, \bibinfo{person}{Zihao Wang},
  \bibinfo{person}{Zihui Gu}, \bibinfo{person}{Zijia Zhu},
  \bibinfo{person}{Zilin Li}, \bibinfo{person}{Zipeng Zhang},
  \bibinfo{person}{Ziwei Xie}, \bibinfo{person}{Ziyi Gao},
  \bibinfo{person}{Zizheng Pan}, \bibinfo{person}{Zongqing Yao},
  \bibinfo{person}{Bei Feng}, \bibinfo{person}{Hui Li}, \bibinfo{person}{J.~L.
  Cai}, \bibinfo{person}{Jiaqi Ni}, \bibinfo{person}{Lei Xu},
  \bibinfo{person}{Meng Li}, \bibinfo{person}{Ning Tian},
  \bibinfo{person}{R.~J. Chen}, \bibinfo{person}{R.~L. Jin},
  \bibinfo{person}{S.~S. Li}, \bibinfo{person}{Shuang Zhou},
  \bibinfo{person}{Tianyu Sun}, \bibinfo{person}{X.~Q. Li},
  \bibinfo{person}{Xiangyue Jin}, \bibinfo{person}{Xiaojin Shen},
  \bibinfo{person}{Xiaosha Chen}, \bibinfo{person}{Xinnan Song},
  \bibinfo{person}{Xinyi Zhou}, \bibinfo{person}{Y.~X. Zhu},
  \bibinfo{person}{Yanping Huang}, \bibinfo{person}{Yaohui Li},
  \bibinfo{person}{Yi Zheng}, \bibinfo{person}{Yuchen Zhu},
  \bibinfo{person}{Yunxian Ma}, \bibinfo{person}{Zhen Huang},
  \bibinfo{person}{Zhipeng Xu}, \bibinfo{person}{Zhongyu Zhang},
  \bibinfo{person}{Dongjie Ji}, \bibinfo{person}{Jian Liang},
  \bibinfo{person}{Jianzhong Guo}, \bibinfo{person}{Jin Chen},
  \bibinfo{person}{Leyi Xia}, \bibinfo{person}{Miaojun Wang},
  \bibinfo{person}{Mingming Li}, \bibinfo{person}{Peng Zhang},
  \bibinfo{person}{Ruyi Chen}, \bibinfo{person}{Shangmian Sun},
  \bibinfo{person}{Shaoqing Wu}, \bibinfo{person}{Shengfeng Ye},
  \bibinfo{person}{T. Wang}, \bibinfo{person}{W.~L. Xiao}, \bibinfo{person}{Wei
  An}, \bibinfo{person}{Xianzu Wang}, \bibinfo{person}{Xiaowen Sun},
  \bibinfo{person}{Xiaoxiang Wang}, \bibinfo{person}{Ying Tang},
  \bibinfo{person}{Yukun Zha}, \bibinfo{person}{Zekai Zhang},
  \bibinfo{person}{Zhe Ju}, \bibinfo{person}{Zhen Zhang}, {and}
  \bibinfo{person}{Zihua Qu}.} \bibinfo{year}{2025}\natexlab{}.
\newblock \bibinfo{title}{DeepSeek-V3.2: Pushing the Frontier of Open Large
  Language Models}.
\newblock
\showeprint[arxiv]{2512.02556}~[cs.CL]
\urldef\tempurl%
\url{https://arxiv.org/abs/2512.02556}
\showURL{%
\tempurl}


\bibitem[Dosovitskiy et~al\mbox{.}(2021)]%
        {vit}
\bibfield{author}{\bibinfo{person}{Alexey Dosovitskiy}, \bibinfo{person}{Lucas
  Beyer}, \bibinfo{person}{Alexander Kolesnikov}, \bibinfo{person}{Dirk
  Weissenborn}, \bibinfo{person}{Xiaohua Zhai}, \bibinfo{person}{Thomas
  Unterthiner}, \bibinfo{person}{Mostafa Dehghani}, \bibinfo{person}{Matthias
  Minderer}, \bibinfo{person}{Georg Heigold}, \bibinfo{person}{Sylvain Gelly},
  \bibinfo{person}{Jakob Uszkoreit}, {and} \bibinfo{person}{Neil Houlsby}.}
  \bibinfo{year}{2021}\natexlab{}.
\newblock \bibinfo{title}{An Image is Worth 16x16 Words: Transformers for Image
  Recognition at Scale}.
\newblock
\showeprint[arxiv]{2010.11929}~[cs.CV]
\urldef\tempurl%
\url{https://arxiv.org/abs/2010.11929}
\showURL{%
\tempurl}


\bibitem[GLM-5-Team et~al\mbox{.}(2026)]%
        {glm-5}
\bibfield{author}{\bibinfo{person}{GLM-5-Team}, \bibinfo{person}{:},
  \bibinfo{person}{Aohan Zeng}, \bibinfo{person}{Xin Lv},
  \bibinfo{person}{Zhenyu Hou}, \bibinfo{person}{Zhengxiao Du},
  \bibinfo{person}{Qinkai Zheng}, \bibinfo{person}{Bin Chen},
  \bibinfo{person}{Da Yin}, \bibinfo{person}{Chendi Ge},
  \bibinfo{person}{Chenghua Huang}, \bibinfo{person}{Chengxing Xie},
  \bibinfo{person}{Chenzheng Zhu}, \bibinfo{person}{Congfeng Yin},
  \bibinfo{person}{Cunxiang Wang}, \bibinfo{person}{Gengzheng Pan},
  \bibinfo{person}{Hao Zeng}, \bibinfo{person}{Haoke Zhang},
  \bibinfo{person}{Haoran Wang}, \bibinfo{person}{Huilong Chen},
  \bibinfo{person}{Jiajie Zhang}, \bibinfo{person}{Jian Jiao},
  \bibinfo{person}{Jiaqi Guo}, \bibinfo{person}{Jingsen Wang},
  \bibinfo{person}{Jingzhao Du}, \bibinfo{person}{Jinzhu Wu},
  \bibinfo{person}{Kedong Wang}, \bibinfo{person}{Lei Li}, \bibinfo{person}{Lin
  Fan}, \bibinfo{person}{Lucen Zhong}, \bibinfo{person}{Mingdao Liu},
  \bibinfo{person}{Mingming Zhao}, \bibinfo{person}{Pengfan Du},
  \bibinfo{person}{Qian Dong}, \bibinfo{person}{Rui Lu},
  \bibinfo{person}{Shuang-Li}, \bibinfo{person}{Shulin Cao},
  \bibinfo{person}{Song Liu}, \bibinfo{person}{Ting Jiang},
  \bibinfo{person}{Xiaodong Chen}, \bibinfo{person}{Xiaohan Zhang},
  \bibinfo{person}{Xuancheng Huang}, \bibinfo{person}{Xuezhen Dong},
  \bibinfo{person}{Yabo Xu}, \bibinfo{person}{Yao Wei}, \bibinfo{person}{Yifan
  An}, \bibinfo{person}{Yilin Niu}, \bibinfo{person}{Yitong Zhu},
  \bibinfo{person}{Yuanhao Wen}, \bibinfo{person}{Yukuo Cen},
  \bibinfo{person}{Yushi Bai}, \bibinfo{person}{Zhongpei Qiao},
  \bibinfo{person}{Zihan Wang}, \bibinfo{person}{Zikang Wang},
  \bibinfo{person}{Zilin Zhu}, \bibinfo{person}{Ziqiang Liu},
  \bibinfo{person}{Zixuan Li}, \bibinfo{person}{Bojie Wang},
  \bibinfo{person}{Bosi Wen}, \bibinfo{person}{Can Huang},
  \bibinfo{person}{Changpeng Cai}, \bibinfo{person}{Chao Yu},
  \bibinfo{person}{Chen Li}, \bibinfo{person}{Chengwei Hu},
  \bibinfo{person}{Chenhui Zhang}, \bibinfo{person}{Dan Zhang},
  \bibinfo{person}{Daoyan Lin}, \bibinfo{person}{Dayong Yang},
  \bibinfo{person}{Di Wang}, \bibinfo{person}{Ding Ai}, \bibinfo{person}{Erle
  Zhu}, \bibinfo{person}{Fangzhou Yi}, \bibinfo{person}{Feiyu Chen},
  \bibinfo{person}{Guohong Wen}, \bibinfo{person}{Hailong Sun},
  \bibinfo{person}{Haisha Zhao}, \bibinfo{person}{Haiyi Hu},
  \bibinfo{person}{Hanchen Zhang}, \bibinfo{person}{Hanrui Liu},
  \bibinfo{person}{Hanyu Zhang}, \bibinfo{person}{Hao Peng},
  \bibinfo{person}{Hao Tai}, \bibinfo{person}{Haobo Zhang}, \bibinfo{person}{He
  Liu}, \bibinfo{person}{Hongwei Wang}, \bibinfo{person}{Hongxi Yan},
  \bibinfo{person}{Hongyu Ge}, \bibinfo{person}{Huan Liu},
  \bibinfo{person}{Huanpeng Chu}, \bibinfo{person}{Jia'ni Zhao},
  \bibinfo{person}{Jiachen Wang}, \bibinfo{person}{Jiajing Zhao},
  \bibinfo{person}{Jiamin Ren}, \bibinfo{person}{Jiapeng Wang},
  \bibinfo{person}{Jiaxin Zhang}, \bibinfo{person}{Jiayi Gui},
  \bibinfo{person}{Jiayue Zhao}, \bibinfo{person}{Jijie Li},
  \bibinfo{person}{Jing An}, \bibinfo{person}{Jing Li},
  \bibinfo{person}{Jingwei Yuan}, \bibinfo{person}{Jinhua Du},
  \bibinfo{person}{Jinxin Liu}, \bibinfo{person}{Junkai Zhi},
  \bibinfo{person}{Junwen Duan}, \bibinfo{person}{Kaiyue Zhou},
  \bibinfo{person}{Kangjian Wei}, \bibinfo{person}{Ke Wang},
  \bibinfo{person}{Keyun Luo}, \bibinfo{person}{Laiqiang Zhang},
  \bibinfo{person}{Leigang Sha}, \bibinfo{person}{Liang Xu},
  \bibinfo{person}{Lindong Wu}, \bibinfo{person}{Lintao Ding},
  \bibinfo{person}{Lu Chen}, \bibinfo{person}{Minghao Li},
  \bibinfo{person}{Nianyi Lin}, \bibinfo{person}{Pan Ta},
  \bibinfo{person}{Qiang Zou}, \bibinfo{person}{Rongjun Song},
  \bibinfo{person}{Ruiqi Yang}, \bibinfo{person}{Shangqing Tu},
  \bibinfo{person}{Shangtong Yang}, \bibinfo{person}{Shaoxiang Wu},
  \bibinfo{person}{Shengyan Zhang}, \bibinfo{person}{Shijie Li},
  \bibinfo{person}{Shuang Li}, \bibinfo{person}{Shuyi Fan},
  \bibinfo{person}{Wei Qin}, \bibinfo{person}{Wei Tian},
  \bibinfo{person}{Weining Zhang}, \bibinfo{person}{Wenbo Yu},
  \bibinfo{person}{Wenjie Liang}, \bibinfo{person}{Xiang Kuang},
  \bibinfo{person}{Xiangmeng Cheng}, \bibinfo{person}{Xiangyang Li},
  \bibinfo{person}{Xiaoquan Yan}, \bibinfo{person}{Xiaowei Hu},
  \bibinfo{person}{Xiaoying Ling}, \bibinfo{person}{Xing Fan},
  \bibinfo{person}{Xingye Xia}, \bibinfo{person}{Xinyuan Zhang},
  \bibinfo{person}{Xinze Zhang}, \bibinfo{person}{Xirui Pan},
  \bibinfo{person}{Xu Zou}, \bibinfo{person}{Xunkai Zhang},
  \bibinfo{person}{Yadi Liu}, \bibinfo{person}{Yandong Wu},
  \bibinfo{person}{Yanfu Li}, \bibinfo{person}{Yidong Wang},
  \bibinfo{person}{Yifan Zhu}, \bibinfo{person}{Yijun Tan},
  \bibinfo{person}{Yilin Zhou}, \bibinfo{person}{Yiming Pan},
  \bibinfo{person}{Ying Zhang}, \bibinfo{person}{Yinpei Su},
  \bibinfo{person}{Yipeng Geng}, \bibinfo{person}{Yong Yan},
  \bibinfo{person}{Yonglin Tan}, \bibinfo{person}{Yuean Bi},
  \bibinfo{person}{Yuhan Shen}, \bibinfo{person}{Yuhao Yang},
  \bibinfo{person}{Yujiang Li}, \bibinfo{person}{Yunan Liu},
  \bibinfo{person}{Yunqing Wang}, \bibinfo{person}{Yuntao Li},
  \bibinfo{person}{Yurong Wu}, \bibinfo{person}{Yutao Zhang},
  \bibinfo{person}{Yuxi Duan}, \bibinfo{person}{Yuxuan Zhang},
  \bibinfo{person}{Zezhen Liu}, \bibinfo{person}{Zhengtao Jiang},
  \bibinfo{person}{Zhenhe Yan}, \bibinfo{person}{Zheyu Zhang},
  \bibinfo{person}{Zhixiang Wei}, \bibinfo{person}{Zhuo Chen},
  \bibinfo{person}{Zhuoer Feng}, \bibinfo{person}{Zijun Yao},
  \bibinfo{person}{Ziwei Chai}, \bibinfo{person}{Ziyuan Wang},
  \bibinfo{person}{Zuzhou Zhang}, \bibinfo{person}{Bin Xu},
  \bibinfo{person}{Minlie Huang}, \bibinfo{person}{Hongning Wang},
  \bibinfo{person}{Juanzi Li}, \bibinfo{person}{Yuxiao Dong}, {and}
  \bibinfo{person}{Jie Tang}.} \bibinfo{year}{2026}\natexlab{}.
\newblock \bibinfo{title}{GLM-5: from Vibe Coding to Agentic Engineering}.
\newblock
\showeprint[arxiv]{2602.15763}~[cs.LG]
\urldef\tempurl%
\url{https://arxiv.org/abs/2602.15763}
\showURL{%
\tempurl}


\bibitem[Jo et~al\mbox{.}(2026)]%
        {jo2026tracediffusionmodelsecretly}
\bibfield{author}{\bibinfo{person}{Sanghyun Jo}, \bibinfo{person}{Ziseok Lee},
  \bibinfo{person}{Wooyeol Lee}, \bibinfo{person}{Jonghyun Choi},
  \bibinfo{person}{Jaesik Park}, {and} \bibinfo{person}{Kyungsu Kim}.}
  \bibinfo{year}{2026}\natexlab{}.
\newblock \bibinfo{title}{TRACE: Your Diffusion Model is Secretly an Instance
  Edge Detector}.
\newblock
\showeprint[arxiv]{2503.07982}~[cs.CV]
\urldef\tempurl%
\url{https://arxiv.org/abs/2503.07982}
\showURL{%
\tempurl}


\bibitem[Lim et~al\mbox{.}(2025)]%
        {lim2025conceptsplitdecoupledmulticonceptpersonalization}
\bibfield{author}{\bibinfo{person}{Habin Lim}, \bibinfo{person}{Yeongseob Won},
  \bibinfo{person}{Juwon Seo}, {and} \bibinfo{person}{Gyeong-Moon Park}.}
  \bibinfo{year}{2025}\natexlab{}.
\newblock \bibinfo{title}{ConceptSplit: Decoupled Multi-Concept Personalization
  of Diffusion Models via Token-wise Adaptation and Attention Disentanglement}.
\newblock
\showeprint[arxiv]{2510.04668}~[cs.CV]
\urldef\tempurl%
\url{https://arxiv.org/abs/2510.04668}
\showURL{%
\tempurl}


\bibitem[Milakov and Gimelshein(2018)]%
        {online-softmax}
\bibfield{author}{\bibinfo{person}{Maxim Milakov} {and}
  \bibinfo{person}{Natalia Gimelshein}.} \bibinfo{year}{2018}\natexlab{}.
\newblock \showarticletitle{Online normalizer calculation for softmax}.
\newblock
\urldef\tempurl%
\url{https://arxiv.org/abs/1805.02867}
\showURL{%
\tempurl}


\bibitem[Paszke et~al\mbox{.}(2019)]%
        {pytorch}
\bibfield{author}{\bibinfo{person}{Adam Paszke}, \bibinfo{person}{Sam Gross},
  \bibinfo{person}{Francisco Massa}, \bibinfo{person}{Adam Lerer},
  \bibinfo{person}{James Bradbury}, \bibinfo{person}{Gregory Chanan},
  \bibinfo{person}{Trevor Killeen}, \bibinfo{person}{Zeming Lin},
  \bibinfo{person}{Natalia Gimelshein}, \bibinfo{person}{Luca Antiga},
  \bibinfo{person}{Alban Desmaison}, \bibinfo{person}{Andreas Köpf},
  \bibinfo{person}{Edward Yang}, \bibinfo{person}{Zach DeVito},
  \bibinfo{person}{Martin Raison}, \bibinfo{person}{Alykhan Tejani},
  \bibinfo{person}{Sasank Chilamkurthy}, \bibinfo{person}{Benoit Steiner},
  \bibinfo{person}{Lu Fang}, \bibinfo{person}{Junjie Bai}, {and}
  \bibinfo{person}{Soumith Chintala}.} \bibinfo{year}{2019}\natexlab{}.
\newblock \bibinfo{title}{PyTorch: An Imperative Style, High-Performance Deep
  Learning Library}.
\newblock
\showeprint[arxiv]{1912.01703}~[cs.LG]
\urldef\tempurl%
\url{https://arxiv.org/abs/1912.01703}
\showURL{%
\tempurl}


\bibitem[Peebles and Xie(2023)]%
        {dit}
\bibfield{author}{\bibinfo{person}{William Peebles} {and}
  \bibinfo{person}{Saining Xie}.} \bibinfo{year}{2023}\natexlab{}.
\newblock \bibinfo{title}{Scalable Diffusion Models with Transformers}.
\newblock
\showeprint[arxiv]{2212.09748}~[cs.CV]
\urldef\tempurl%
\url{https://arxiv.org/abs/2212.09748}
\showURL{%
\tempurl}


\bibitem[Tian et~al\mbox{.}(2024)]%
        {tian2024diffuseattendsegmentunsupervised}
\bibfield{author}{\bibinfo{person}{Junjiao Tian}, \bibinfo{person}{Lavisha
  Aggarwal}, \bibinfo{person}{Andrea Colaco}, \bibinfo{person}{Zsolt Kira},
  {and} \bibinfo{person}{Mar Gonzalez-Franco}.}
  \bibinfo{year}{2024}\natexlab{}.
\newblock \bibinfo{title}{Diffuse, Attend, and Segment: Unsupervised Zero-Shot
  Segmentation using Stable Diffusion}.
\newblock
\showeprint[arxiv]{2308.12469}~[cs.CV]
\urldef\tempurl%
\url{https://arxiv.org/abs/2308.12469}
\showURL{%
\tempurl}


\bibitem[Tillet et~al\mbox{.}(2019)]%
        {triton}
\bibfield{author}{\bibinfo{person}{Philippe Tillet}, \bibinfo{person}{H.~T.
  Kung}, {and} \bibinfo{person}{David Cox}.} \bibinfo{year}{2019}\natexlab{}.
\newblock \showarticletitle{Triton: an intermediate language and compiler for
  tiled neural network computations}. In \bibinfo{booktitle}{\emph{Proceedings
  of the 3rd ACM SIGPLAN International Workshop on Machine Learning and
  Programming Languages}} (Phoenix, AZ, USA) \emph{(\bibinfo{series}{MAPL
  2019})}. \bibinfo{publisher}{Association for Computing Machinery},
  \bibinfo{address}{New York, NY, USA}, \bibinfo{pages}{10–19}.
\newblock
\showISBNx{9781450367196}
\href{https://doi.org/10.1145/3315508.3329973}{doi:\nolinkurl{10.1145/3315508.3329973}}


\bibitem[Vaswani et~al\mbox{.}(2023)]%
        {transformer}
\bibfield{author}{\bibinfo{person}{Ashish Vaswani}, \bibinfo{person}{Noam
  Shazeer}, \bibinfo{person}{Niki Parmar}, \bibinfo{person}{Jakob Uszkoreit},
  \bibinfo{person}{Llion Jones}, \bibinfo{person}{Aidan~N. Gomez},
  \bibinfo{person}{Lukasz Kaiser}, {and} \bibinfo{person}{Illia Polosukhin}.}
  \bibinfo{year}{2023}\natexlab{}.
\newblock \bibinfo{title}{Attention Is All You Need}.
\newblock
\showeprint[arxiv]{1706.03762}~[cs.CL]
\urldef\tempurl%
\url{https://arxiv.org/abs/1706.03762}
\showURL{%
\tempurl}


\bibitem[Wang et~al\mbox{.}(2020)]%
        {minilm}
\bibfield{author}{\bibinfo{person}{Wenhui Wang}, \bibinfo{person}{Furu Wei},
  \bibinfo{person}{Li Dong}, \bibinfo{person}{Hangbo Bao}, \bibinfo{person}{Nan
  Yang}, {and} \bibinfo{person}{Ming Zhou}.} \bibinfo{year}{2020}\natexlab{}.
\newblock \bibinfo{title}{MiniLM: Deep Self-Attention Distillation for
  Task-Agnostic Compression of Pre-Trained Transformers}.
\newblock
\showeprint[arxiv]{2002.10957}~[cs.CL]
\urldef\tempurl%
\url{https://arxiv.org/abs/2002.10957}
\showURL{%
\tempurl}


\bibitem[Xiong et~al\mbox{.}(2026)]%
        {D2Prune}
\bibfield{author}{\bibinfo{person}{Lang Xiong}, \bibinfo{person}{Ning Liu},
  \bibinfo{person}{Ao Ren}, \bibinfo{person}{Yuheng Bai},
  \bibinfo{person}{Haining Fang}, \bibinfo{person}{Binyan Zhang},
  \bibinfo{person}{Zhe Jiang}, \bibinfo{person}{Yujuan Tan}, {and}
  \bibinfo{person}{Duo Liu}.} \bibinfo{year}{2026}\natexlab{}.
\newblock \showarticletitle{D2 Prune: Sparsifying Large Language Models via
  Dual Taylor Expansion and Attention Distribution Awareness}.
\newblock \bibinfo{journal}{\emph{Proceedings of the AAAI Conference on
  Artificial Intelligence}} \bibinfo{volume}{40}, \bibinfo{number}{32}
  (\bibinfo{date}{March} \bibinfo{year}{2026}), \bibinfo{pages}{27171–27179}.
\newblock
\showISSN{2159-5399}
\href{https://doi.org/10.1609/aaai.v40i32.39932}{doi:\nolinkurl{10.1609/aaai.v40i32.39932}}


\bibitem[Yang and Zhang(2024)]%
        {yang2024fla}
\bibfield{author}{\bibinfo{person}{Songlin Yang} {and} \bibinfo{person}{Yu
  Zhang}.} \bibinfo{year}{2024}\natexlab{}.
\newblock \bibinfo{booktitle}{\emph{FLA: A Triton-Based Library for
  Hardware-Efficient Implementations of Linear Attention Mechanism}}.
\newblock
\urldef\tempurl%
\url{https://github.com/fla-org/flash-linear-attention}
\showURL{%
\tempurl}


\bibitem[Zadouri et~al\mbox{.}(2026)]%
        {flashattention4}
\bibfield{author}{\bibinfo{person}{Ted Zadouri}, \bibinfo{person}{Markus
  Hoehnerbach}, \bibinfo{person}{Jay Shah}, \bibinfo{person}{Timmy Liu},
  \bibinfo{person}{Vijay Thakkar}, {and} \bibinfo{person}{Tri Dao}.}
  \bibinfo{year}{2026}\natexlab{}.
\newblock \bibinfo{title}{FlashAttention-4: Algorithm and Kernel Pipelining
  Co-Design for Asymmetric Hardware Scaling}.
\newblock
\showeprint[arxiv]{2603.05451}~[cs.CL]
\urldef\tempurl%
\url{https://arxiv.org/abs/2603.05451}
\showURL{%
\tempurl}


\end{thebibliography}

\end{document}